\documentclass[final,3p,times]{elsarticle}

\usepackage{amssymb}
\usepackage{xcolor}
\usepackage{multirow}
\usepackage{natbib}
\usepackage{amsmath}

\usepackage{lineno}

\journal{Computer Physics Communications}

\begin{document}

\begin{frontmatter}



\title{PINTO: Physics-informed transformer neural operator for learning generalized solutions of partial differential equations for any initial and boundary condition}


\author[1]{Sumanth Kumar Boya}
\author[1]{Deepak N. Subramani\corref{cor1}%
}
\ead{deepakns@iisc.ac.in}
\cortext[cor1]{Corresponding author}

\address[1]{Department of Computational and Data Sciences, Indian Institute of Science, Bangalore 560012 India }


\begin{abstract}
Applications in physics, engineering, mechanics, and fluid dynamics necessitate solving nonlinear partial differential equations (PDEs) with different initial and boundary conditions. Operator learning, an emerging field, solves these PDEs by employing neural networks to map the infinite-dimensional input and output function spaces. These neural operators are trained using data (observations or simulations) and PDE residuals (physics loss). A key limitation of current neural methods is the need to retrain for new initial/boundary conditions and the substantial simulation data required for training. We introduce a physics-informed transformer neural operator (named PINTO) that generalizes efficiently to new conditions, trained solely with physics loss in a simulation-free setting. Our core innovation is the development of iterative kernel integral operator units that use cross-attention to transform domain points of PDE solutions into initial/boundary condition-aware representation vectors, supporting efficient and generalizable learning. The working of PINTO is demonstrated by simulating important 1D and 2D equations used in fluid mechanics, physics and engineering applications: advection, Burgers, and steady and unsteady Navier-Stokes equations (three flow scenarios). We show that under challenging unseen conditions, the relative errors compared to analytical or numerical (finite difference and volume) solutions are low, merely 20\% to 33\% of those obtained by other leading physics-informed neural operator methods. Furthermore, PINTO accurately solves advection and Burgers equations at time steps not present in the training points, an ability absent for other neural operators. The code is accessible at \texttt{https://github.com/quest-lab-iisc/PINTO}.

\end{abstract}


\begin{highlights}
\item New iterative integral kernel neural operator for learning generalized PDE solutions
\item Cross attention units transform solution domain representation to be boundary aware
\item Lifting and projection layers to map function spaces for efficient representation
\item Five challenging test cases demonstrate utility and improvement over leading models
\item Low relative error and ability to find solutions outside training examples
\end{highlights}

\begin{keyword}


Nonlinear mapping \sep Operator learning \sep Transformer operator \sep Scientific machine learning \sep Physics-informed deep learning \sep Initial and boundary generalization \sep AI for PDEs

\end{keyword}

\end{frontmatter}


\section{Introduction}

Solving partial differential equations (PDEs) with different initial and boundary conditions is essential to understand and characterize the dynamics of natural and engineering systems. Traditionally, numerical methods (e.g., finite difference, finite element, finite volume, or spectral) are employed to solve PDEs. In recent decades, scientists have created high-performance computing software designed to solve PDEs, enabling their use as forward models in engineering design \citep{wang2022tensorflow, ion2022tensor, richter2021solving, heermann2012parallel}. However, for a new set of initial and/or boundary conditions, the numerical method must be applied afresh to solve the PDE as a new problem by spending significant compute resources. In contrast, contemporary artificial intelligence tasks in computer vision and natural language processing involve running a trained model (inference), which is computationally less expensive, for new scenarios. Those fields have benefitted from neural models that generalize to new situations well. The promise and application scope of such PDE solving models that could be trained once and used for fast simulation in new situations is appealing. Thus, over the past several years, the development of neural models for simulating systems governed by PDEs has picked up pace. Our goal is to develop a neural model for learning generalized solutions of PDEs for any initial and boundary condition.


Deep neural networks to solve PDEs are broadly classified into two categories: \textit{(i)} neural operators that learn the mapping between function spaces (e.g., input function of say parameters and output function of the solution) such as DeepONets, Fourier neural operators (FNO), physics-informed neural operators (PINO), and graph neural operators (GNO) \citep{li2020neural, li2020fourier,  lu2021learning, you2022nonlocal, kovachki2023neural} and \textit{(ii)} neural networks that learn a map between the PDE's space-time domain and the solution function \citep{RAISSI2019686, lu2021deepxde, karniadakis2021physics, yang2021b, cai2021physics}. This distinction underscores the primary character of the input and output spaces used in these approaches. The latter is further classified into three \textit{(i)} Physics-Informed Neural Networks (PINNs), \textit{(ii)} Physics-Guided Neural Networks (PGNN) and \textit{(iii)} Physics-Encoded Neural Networks (PENNs) \citep{faroughi2024physics}. 

All approaches can be trained using a combination of simulation data and PDE-residuals. The PGNN class of neural networks is trained primarily using simulations. They require a large data set of simulated (e.g., by solving PDEs numerically using traditional methods) or observed (experimental data) input and output. The PINN class of neural networks imposes the underlying physics by modifying the training objective to include the residuals of the PDEs. The residuals are calculated through automatic differentiation or through the discrete form of the PDEs in either weak or strong form \citep{yamazaki2025finite, kaewnuratchadasorn2024physics}. Energy-based loss functions have also been proposed \citep{nguyen2020deep, abueidda2022deep, he2023deep}. The PENN class of neural networks attempts to impose known physics as part of the architecture itself, such as making the activation function learnable. 


PGNN, PINN and PENN learn the map for one instance of the PDE and require re-training for solving with new initial/boundary conditions, much like traditional numerical solvers. On the other hand, neural operators can learn a map between an initial/boundary function and a solution function, thereby achieving generalization to unseen conditions for a particular PDE. However, many of the current leading neural operators such as the Fourier neural operator (FNO) \citep{li2020fourier, pathak2022fourcastnet, wen2022u, bonev2023spherical, you2022learning, li2023fourier,  george2024incrementalspatialspectrallearning, kovachki2021universal, JMLR:v24:21-1524, lehmann20243d}, DeepONet \citep{wang2021learning, lu2021learning, lanthaler2022error, he2024sequential, xu2023transfer, li2023phase, goswami2022physics, goswami2023physics, goswami2024learning, haghighat2024deeponet, kobayashi2024improved, zhang2024energy, he2024geom, koric2023data}, Graph neural operator \citep{li2020neural, li2020multipole}, Convolutional neural operators \citep{raonic2024convolutional}, Complex neural operator \citep{tiwari2023cono}, Wavelet neural operator \citep{gupta2021multiwavelet, tripura2023wavelet, rani2023fault}, Laplacian neural operator \citep{pang2020npinns, cao2024laplace}, RiemannONets \citep{peyvan2024riemannonets}, Geometry informed neural operator (GINO) \citep{li2024geometry}, Diffeomorphism Neural Operator \citep{zhao2024diffeomorphism}, Spectral neural operator \citep{fanaskov2023spectral, rafiq2022ssno}, operator transformer \citep{hao2023gnot, li2022transformer, shih2025transformers}, Local neural operator \cite{li2024local}, and Peridynamic neural operators \cite{jafarzadeh2024peridynamic} are all trained using a small to large amount of simulation data. They resemble the PGNN framework; however, due to the fundamental difference in the input-output spaces, neural operators are generalizable. Some of the above neural operators have a physics-informed variant \citep{li2024physics, wang2021learning}. However, these also require some amount of simulation data (perhaps in a coarse domain) for training. Moreover, as we show in this paper, the current leading architectures that can be trained without any simulation data are not efficient in practice for initial/boundary condition generalization. 


For learning generalized solutions of PDEs, we develop the Physics-Informed Neural Transformer Operator (PINTO) that learns a map between the initial/boundary function space and the PDE's solution function space. The primary novelty is introducing a cross-attention unit that modulates the internal representation of the domain's space-time coordinates with information from the initial/boundary function, so that a generalized solution is obtained for any unseen initial/boundary function. Our model is trained with physics loss computed by automatic differentiation, without using simulation data. 

The applications that can benefit from the developed architecture are vast. For example, it could be used as surrogate models for faster inference \citep{chennault2021adjoint, arcucci2020neural} and as solvers \citep{wu2021fast, fablet2021learning} in data assimilation and inverse problems. It can replace traditional CFD solvers in applications that require fluid flow simulations such as autonomous path planning \citep{chowdhury2023intelligent}, biomedical fluid dynamics \citep{kelley_thomas_2023}, mask design \citep{mittal_et_al2023}, and wake modeling in wind farms \citep{breton_et_al2017, gafoor2025physics}.

\subsection{Related work:}\label{sec: literature_survey} 

Neural operators to learn maps between infinite-dimensional input and output spaces have gained tremendous popularity in recent years. Concurrent work on branch-trunk network DeepONet architecture, discretization-dependent convolutional neural operators, and discretization-invariant Fourier neural operators has led the way. In the DeepONet architecture, a trunk net encodes the coordinates of the space-time domain of the PDE, and a branch net encodes the discrete input function space \citep{lu2021learning}. The representations are merged using a dot product. This operator was trained using simulation data. Later work extended the setup to situations where data availability is sparse by introducing PDE-residual loss as a physics-informed regularizer during DeepONet training \citep{wang2021learning, goswami2023physics}. Extensions have been reported to multiple input functions \citep{jin2022mionet}, efficient numerics \citep{mandl2024separable}, and training using the variational form of PDE residuals \cite{goswami2022physics}. 

The Fourier neural operator (FNO) is a discretization invariant neural operator that is emerging as a potential universal approximator \citep{li2020fourier,kovachki2021universal,kovachki2023neural}. Using convolutions within the spectral domain facilitated by the Fast Fourier Transform, FNOs learn the mapping between function spaces. However, they require a substantial amount of simulation data for training purposes. Enhancements in FNOs have been made by incorporating PDE-residual loss in their training, although they still require simulation data, but with reduced resolution \citep{li2024physics}. Deep implicit FNO has been used to learn the mapping between loading conditions and the response of heterogeneous materials without predefined constitutive models or microstructure measurements \citep{you2022learning}. An FNO-based architecture has been developed to simulate a highly complex multiphase $\text{CO}_{2}$ water problem with wide ranges of porosity heterogeneity, anisotropy, reservoir conditions, injection configurations, flow rates, and multiphase flow properties using large simulation datasets \cite{wen2022u}. Fourier neural operator has been used for full wave inversion to infer subsurface structure information from seismic waveform data \cite{zhu2023fourier}.

Transformers have been suggested as neural operators because they are inherently capable of mapping functional spaces over irregular and nonuniform grids. Data-driven operator learning using an attention mechanism reminiscent of Galerkin methods has been proposed \cite{NEURIPS2021_d0921d44}. With this Galerkin-like attention, an encoder-decoder transformer model was developed to handle both uniform and varied discretization grids \cite{li2022transformer}. To solve complex PDEs involving multiple input functions and irregular grids, a general neural operator transformer \cite{hao2023gnot} was introduced. The authors implemented a heterogeneous normalized attention mechanism to decrease the computational cost of calculating attention scores and introduced a geometric gating mechanism to address multiscale problems. 


Recently, several approaches have been proposed to solve PDEs without using simulation data. For example, a finite element-based physics-informed operator learning framework has been developed with a Galerkin discretized weak formulation and an implicit Euler time integration scheme for temporal discretization as a loss function to incorporate physics into the network \cite{yamazaki2025finite}. Similarly, the Lippmann-Schwinger operator in Fourier space has been used to construct physical constraints, effectively eliminating Automatic Differentiation \citep{harandi2024spectral}.  The prior knowledge of the governing equations has been encoded through a stochastic projection-based gradient \cite{navaneeth2023stochastic} to solve PDEs for multiple initial and boundary conditions without retraining the model. Still, these approaches come with the cost of approximation and randomness in evaluating gradients.


Though DeepONet achieved a wide range of applicability in approximating nonlinear operators, it has issues with scalability for high-dimensional PDEs due to data requirements. Also, the discrete input function requires data on a pre-defined grid that makes generalization difficult. FNOs and other recent neural operators (see Section 1) cannot be trained reliably without simulation data, restricting their applications to users without access to simulation data. The attention mechanism is a promising route for developing neural operators, and we utilize the same in our work. As a reference, we used the solution obtained by a physics-informed DeepONet (PI-DeepONet) to benchmark the performance of our proposed model.

\subsection{Key contributions}\label{sec: key_contribution} 

Our main contributions are: \textit{(i)} development of a new cross-attention mechanism, grounded in neural operator theory, for an efficient neural operator that generalizes the solution of initial boundary value problems for unseen initial/boundary conditions; \textit{(ii)} training of the neural operator uses only physics loss, without data from numerical simulations; \textit{(iii)} better simulation than leading physics informed neural operator models for challenging benchmark 1D and 2D test cases including the advection equation, Burger's equation, and Navier Stokes Equations in three different test cases; and \textit{(iv)} ability to simulate solutions outside the temporal domain used for training. The success of our new neural operator is attributable to the transformation of the PDE solution's domain query point (spatio-temporal coordinates) into a boundary-aware vector through multiple cross-attention units (iterative kernel integral operator).  

In the remaining sections, we first describe the mathematical problem statement, the neural operator theory, and the practical implementation of our new architecture (Sect.~\ref{sec:development}). Next, we apply the new network for the $1$D Advection, $1$D nonlinear Burgers, the Kovasznay flow, the Beltrami flow, and the lid-driven cavity flow test cases (Sect.~\ref{sec:applications}). We compare our results with other leading physics-informed neural operator architectures that can be trained without simulation data (i.e., PI-DeepONet) and showcase the advantages of the new model for initial and boundary condition generalization. In Sect.~\ref{sec:discussion}, we discuss the computational complexity, differnce of PINTO from other physics-informed neural networks and neural operators, how to leverage our new cross-attention unit in other operators and limitations of PINN models. Finally, we conclude with a summary and directions for future work (Sect.~\ref{sec:conclusion}). 

\begin{table}[!ht]
    \centering
    \caption{\textbf{List of symbols with their meaning used in the present paper.}}
    \label{tab: symbollist}
    \begin{tabular}{|l|l|}
        \hline
         $\mathcal{N}$ & Nonlinear differential operator\\
         \hline
         $\mathcal{B}$ & Initial/boundary operator\\
         \hline
         $\mathcal{H}$ & Solution space \\
         \hline 
         $\Omega,\,\partial\Omega$ & Spatio-temporal domain, and domain's boundary\\
         \hline 
         $X,\,X_{b}$ & $d-$dimensional spatio-temporal coordinate in $\Omega$ and $\partial\Omega$ respectively\\
         \hline
         $\mathcal{A}$ & Functional space of initial and boundary conditions\\
         \hline
         $h$ & $s-$dimensional solution field in $\mathcal{H}$\\
         \hline
         $f,\,b$ & Forcing term and Dirichlet boundary conditions respectively\\
         \hline
         $\alpha$ & PDE's parameter vector\\
         \hline
         $\mathcal{G}$ & Map between initial and boundary conditions functional space $\left(\mathcal{A}\right)$ to solution space $\left(\mathcal{H}\right)$\\
         \hline
         $\mathcal{G}_{\theta}$ & Neural operator that approximates $\mathcal{G}$ with $\theta$ as trainable parameters\\
         \hline
         $\Theta,\,\Theta^{*}$ & Trainable parameters $\theta$ and its optimal value\\
         \hline
         $p_n$ & Number of parameters in the neural network (number of elements in $\Theta$)\\
         \hline
         $N_{c}$ & Number of collocation points\\
         \hline
         $N_{ib}$ & Number of initial and boundary points\\
         \hline
         $\lambda_{i}$ & Weights assigned to the $i$th term in the loss function\\
         \hline
         $\mathcal{Q}$ & Projection operator that projects the learned representation from iterative kernel integration to solution space\\
         \hline
         $\mathcal{P}$ & Lifting operator that lifts input to a higher dimensional space\\
         \hline
         $\mathcal{W}_{j}$ & Local linear operator\\
         \hline
         $\mathcal{K}_{j}$ & Nonlinear iterative kernel integral operator\\
         \hline
         $\gamma_{j}$ & Bias term\\
         \hline
         $\sigma$ & Nonlinear activation function\\
         \hline
         $\mu_{j}$ & Hidden representation\\
         \hline
         $D_{j}$ & Domain of hidden representation\\
         \hline
         $A_{h},\,B_{h},\text{ and }R_{h}$ & Linear transformation in an attention head\\
         \hline
         $m$ & Dimension of the encoded vector\\
         \hline
         $\zeta_{i}$ & Attention score corresponds to $i^{th}$ value token\\
         \hline
         $L$ & Length of discretized initial/boundary points \\
         \hline
         $x, y, t$ & Space and time coordinates in the test cases.\\
         \hline
         $u, v, p$ & $x,\,y$-directional velocities, and pressure in Navier Stokes test cases\\
         \hline
         $\beta$ & Constant advection speed in advection test case\\
         \hline
         $\omega_{i},\,a_{i},\,\phi_{i},\,\Delta_{x}$ & wave number, amplitude, phase angle, and domain size\\ 
         &in sinusoidal wave equation for initial conditions of advection test case\\
         \hline
         $\nu$ & Kinematic viscosity in Navier Stokes test cases\\
         \hline 
    \end{tabular}
\end{table}

\section{Development of Physics-Informed Transformer Neural Operator} \label{sec:development}
\subsection{Neural Operator Definition and Loss Function}\label{sec:problem_statement}
Consider a general partial differential equation (PDE) of the form
\begin{subequations}
\begin{align}
    \mathcal{N}\left(h, X; \mathbf{\alpha}\right) = f\,\text{in}\,\Omega,\\
    \mathcal{B}\left(h, X_b \right) = b\,\text{on}\,\partial\Omega,
\end{align}\label{eq:def:pde}
\end{subequations}
where $h\in\mathcal{H}\subseteq \mathbb{R}^s$ is the $s$-dimensional solution field from the solution space $\mathcal{H}\subset L^2(\Omega,\mathbb{R}^s)$, $\mathcal{N}$ is a general nonlinear differential operator with spatial and temporal partial derivatives, $X\in \Omega \subseteq \mathbb{R}^{d}$ is a $d$-dimensional coordinate from the spatiotemporal domain $\Omega$, 
$\mathbf{\alpha}$ is the PDE's parameter vector, $f$ is a forcing term, $\mathcal{B}$ is the initial/boundary operator, $X_b \in \partial\Omega \subseteq \mathbb{R}^d$ is the $d$-dimensional initial/boundary coordinate from the domain's boundary $\partial \Omega$ and $b$ is the imposed initial/boundary condition vector. Table~\ref{tab: symbollist} contains the list of symbols used in this paper with their definitions.

The initial boundary value problem can be expressed as follows. Let $\mathcal{A} \subset L^2(\partial\Omega,\mathbb{R}^d)$ be the functional space of the initial/boundary condition. For an imposed initial/boundary condition $b\in \mathcal{A}$, there is a solution $h\in \mathcal{H}$ that satisfies the PDE \eqref{eq:def:pde}. Thus, there exists a map $\mathcal{G}: \mathcal{A}\rightarrow\mathcal{H}$ that for every observed pair $(b,h)$ satisfies $h=\mathcal{G}(b)$. The solution field is accessible at any $X\in\Omega$ as $h(X)=\mathcal{G}(b)(X)$. We develop a parametrized neural operator $\mathcal{G}_\theta(X,b;\Theta^{*}), \Theta^{*}\in\mathbb{R}^{p_n}$ that approximates $\mathcal{G}$ and generalizes to predict the correct $h(X)$ for any $b\in \mathcal{A}$. Here, ${p_n}$ is the dimension of the parameter vector of the neural network.

We propose the use of PDE-residuals (physics-loss) to train $\mathcal{G}_\theta$. The set of equations that $\mathcal{G}_\theta$ must satisfy is as follows.
\begin{subequations}
    \begin{align}
        \mathcal{N}(\mathcal{G}_\theta(X,b;\Theta^{*}),X;\alpha) = f \text{ on }\Omega,\\
        \mathcal{B}(\mathcal{G}_\theta(X,b;\Theta^{*}),X_b) = b\text{ on }\partial\Omega\, \forall\,b\in\mathcal{A}.
    \end{align}
\end{subequations}
Thus, the training objective to compute the optimal $\Theta^*$ by emperical risk minimization can be expressed as 
\begin{equation}
    \min_{\Theta}\,\sum_{k=1}^{K}\{\frac{\lambda_{1}}{N_{c}}\sum_{j=1}^{N_{c}}|f_{k}^{j}-\mathcal{N}(\mathcal{G}(\Theta;X_{k}^{j});\alpha)|^{2} + \frac{\lambda_{2}}{N_{ib}}\sum_{j=1}^{N_{ib}}|b_{k}^{j}-\mathcal{B}(\mathcal{G}(\Theta;X_{b,\,k}^{j}))|^{2}\}\,,\label{eq:mod_PINNs_loss}
\end{equation}
where $k=[1,2,\cdots,K]$ is the set of discrete boundary conditions sampled from $\mathcal{A}$ such that $b_k\in\mathcal{A}$, $\lambda_{1},\,\lambda_{2}$ are the weights to balance the different terms in the loss function. The first term in the above equation \eqref{eq:mod_PINNs_loss} is the physics loss and the second term forces the neural operator to satisfy the provided boundary conditions. Now the goal is to develop an architecture for $\mathcal{G}_\theta$ that is efficient in learning the representation of the PDE's solution for any initial and boundary condition. In what follows, we develop a neural transformer architecture for the same.

\subsection{Cross Attention Neural Operator Theory}\label{sec:mathematical_theory}
Consider the parametric map $\mathcal{G}_{\Theta}\colon\mathcal{A}\to\mathcal{H}$ as a composition of neural layers that perform lifting, iterative kernel integration, and projection as follows \cite{li2023fourier}.
\begin{equation}
\mathcal{G}_{\theta} := \mathcal{Q} \circ \sigma(\mathcal{W}_{T} + \mathcal{K}_{T} + \gamma_{T}) \circ \cdot \cdot \cdot \circ  \sigma(\mathcal{W}_{1} + \mathcal{K}_{1} + \gamma_{1}) \circ \mathcal{P}\,,\label{eq:compositional_map}    
\end{equation}
where $\mathcal{P}$ is the lifting operator, $\mathcal{Q}$ is the projection operator and the other terms are the $j=1,\cdots, T$ hidden layers that perform iterative kernel integration. Here, $\sigma$ is the pointwise nonlinear activation function, $\mathcal{W}_{j}$ is a local linear operator, $\mathcal{K}_{j}$ is the nonlinear kernel integral operator and $\gamma_{j}$ is the bias function. These layers map each hidden representation to the next, i.e., $\{\mu_{j}\colon D_{j}\to\mathbb{R}^{d_{\mu_{j}}}\}\mapsto \{\mu_{j+1}\colon D_{j+1}\to \mathbb{R}^{d_{\mu_{j+1}}}\}$ using 
\begin{equation}
    \mu_{j+1}(x) = \sigma_{j+1}\left(\mathcal{W}_{j}\mu_{j}(x) + \int_{D_{j}} \left(\kappa^{(j)}(x,y)\mu_{j}(y)\right) \,d\mu_{j}(y) + \gamma_{j}(x)\right) \,\, \forall \,x \in D_{j+1} \label{eq: NO_hidden_layer}\,.
\end{equation}

We propose an encoding layer $\mathcal{P}=MLP_{qpe}(X)$ for the lifting operator and a multi-layer perceptron $\mathcal{Q}=MLP(\mu_T(X))$ for the projection operator. For $\mathcal{K}_j$, we propose a cross-attention kernel integral operator 
\begin{equation}
   \mathcal{K}_j = \int_{\partial\Omega}\frac{\text{exp}\left(\frac{\langle A_{h}\mu_j(X), B_{h}MLP_{bpe}\left(X_{b}\right)\rangle}{\sqrt{m}}\right)}{\int_{\partial\Omega}\text{exp}\left(\frac{\langle A_{h}\mu_j(X), B_{h}MLP_{bpe}\left(X_{b}\right)\rangle}{\sqrt{m}}\right)\rm{d}X_{b}}R_{h}MLP_{bve}\left(b\right)\,\rm{d}X_{b}\,,
\end{equation}
where $\mu_j(X)$ is the hidden representation of the boundary domain's query point coordinate, $MLP_{bpe}$ is the encoding fully connected layers for initial/boundary domain's coordinate $X_b$, $MLP_{bve}$ is the encoding fully connected layers for the boundary condition function value $b$, $A_h$ is the linear transformation for the encoded query point, $B_h$ is the linear transformation for the encoded boundary point and $R_h$ is the linear transformation for the encoded boundary function value, $m$ is the dimension of the encoded vectors (hyperparameter) and $\langle\cdot,\cdot\rangle$ is the Euclidean inner product on $\mathbb{R}^m$.

\subsection{Practical Implementation}\label{sec:framework}
\begin{figure}
\centering
\includegraphics[width=1\columnwidth]{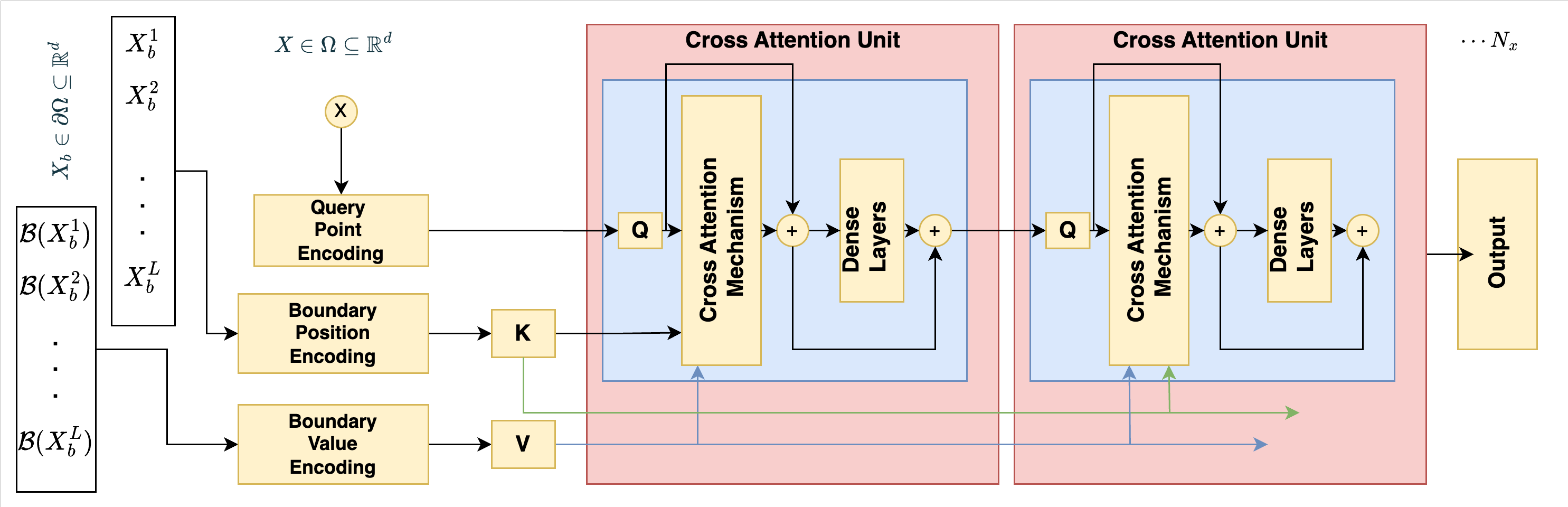}
\caption{\textbf{Schematic of PINTO:} Our neural operator has three stages (i) Query Point, Initial/Boundary Point and Initial/Boundary Value (lifting operators) (ii) Cross Attention Units (iterative kernel intergral operators) and (iii) output projection dense layers.}
\label{fig: neural_schematics}
\end{figure}
Fig.~\ref{fig: neural_schematics} illustrates our new Physics-Informed Neural Transformer Operator, named PINTO, that implements the cross-attention kernel integral operator. There are three stages of operation. In the first stage, we discretize the boundary function $\mathcal{B}(X_b^i)$ at $L$ discrete points $X_b^i$, $i=1,\cdots, L$, and encode them using Boundary Value Encoding (BVE) and Boundary Position Encoding (BPE) units. We also encode the solution domain's query point coordinate using Query Point Encoding (QPE) unit. All these encoding units are lifting operators as they encode scalar values as vectors of dimension $m$.  We use dense layers in these units, but note that any other type of layer such as convolution or recurrent could also be used in the approach. 

In the second stage, the network obtains a boundary-aware query point encoding vector through multiple passes of our new cross-attention units. In each cross-attention unit, the boundary key and value are shared so that the initial/boundary conditions influence the context-aware hidden representation of query points correctly. The vector representation from the QPE is used as the query of the first cross-attention block, and the output of the previous cross-attention unit is used for all subsequent cross-attention units. We implement a deep multihead cross attention unit to increase the generalization ability and learning capacity of the neural operator model. The discrete multihead cross-attention unit is implemented as follows. First, the attention score $\zeta_{i}$ between the encoded query point at the layer $j$ and every $i$th discrete initial/boundary point and value is calculated as
\begin{equation}\label{eq: attention_score}
    \zeta_{i} =  \left(\sum_{l=1}^{L} \text{exp}\left(\frac{\langle A_{h}\mu_j(X) \; , \; B_{h}MLP_{bpe}(X_{b}^{l}) \rangle}{\sqrt{m}}\right) \,\right)^{-1} \text{exp}\left(\frac{\langle A_{h}\mu_j(X),\,B_{h}MLP_{bpe}(X_{b}^{i})\rangle}{\sqrt{m}}\right) \,.
\end{equation}
Next, the output from the cross-attention unit with $H$ attention heads is calculated with a residual connection and a Swish nonlinear activation function as
\begin{equation}\label{eq: attention_output}
    \mu_{j+1}(X) = \sigma\left(A_h\mu_j(X) + \left(\sum_{h=1}^{H} \sum_{i=1}^{L}\zeta_{i}\cdot R_{h}MLP_{bve}(\mathcal{B}(X_{b}^{i}))\right)\right)\,.
\end{equation}
In the third and final stage, the encoded query point representation is projected onto the solution space using dense layers. 

During training, the data must be prepared such that the QPE unit is exposed to collocation points within the domain where the PDE loss is applied, as well as boundary points where both the boundary condition and the PDE loss are enforced. For each input to the QPE unit, the corresponding boundary conditions and the sequence of their values must be provided to the BPE and BVE units. In the case of Dirichlet boundary conditions, the boundary value is provided directly. However, for Neumann boundary conditions, the derivative values need to be fed to the BVE unit, and the loss function needs to be modified to account for these conditions. The code, PINTO architecture, and their trained weights have been released with a Zenodo DOI \cite{b_sumanth_kumar_2024_14330861}.

\section{Applications of PINTO}\label{sec:applications}

The performance of PINTO in simulating the solution of PDEs for unseen initial and boundary conditions is demonstrated by employing the following test cases: \textit{(i)} $1$D-Advection equation, \textit{(ii)} $1$D non-linear Burgers equation, \textit{(iii)} $2$D unsteady Beltrami flow, \textit{(iv)} $2$D steady Kovasznay flow, and \textit{(v)} $2$D steady Lid-driven flow in the $x-y/z$ plane. The first two test cases demonstrate the applicability of our PINTO model in solving initial value problems and providing solutions for unseen times in the temporal domain. The latter ability is absent for the other leading physics-informed neural operators as we show. Numerical solutions of the advection equation from the PDEBENCH dataset \citep{takamoto2022pdebench}, and those of the Burgers equation from an off-the-shelf solver \cite{kovachki2021neural} are used to validate the PINTO simulations. The next three test cases are governed by the Navier-Stokes equations. The third and fourth test cases, namely the Kovasznay and Beltrami flows, have analytical solutions \citep{kovasznay_1948, taylor1923lxxv} for different Reynolds numbers (Re), which are used to validate the PINTO solutions. These test cases demonstrate the applicability of our PINTO model to solve equations where the initial and boundary conditions vary spatially and temporally. In the fifth test case, called lid-driven cavity flow, we investigate the steady state flow within a vertical cavity where the top lid moves to the right at a constant speed. This flow is a simplified version of the wind-driven ocean flow and finds various practical applications in ocean and atmospheric studies. To validate our PINTO predictions, we obtain numerical solutions of the flow from a Finite Volume code \citep{ueckermann_and_lermusiaux_MSEAS2012}, where the boundary condition varied is the speed of the lid movement. Note that in all these situations, the validation data is not used for training. Training is performed using only Physics loss and initial/boundary conditions.

The performance of the PINTO model for all numerical examples is compared to the physics-informed DeepONet model (PI-DeepONet) \citep{lu2021learning,goswami2023physics}. 
PI-DeepONet is chosen as the benchmark for comparison as it can be trained using only physics-loss unlike other neural operators that need some amount of simulation data. Table~\ref{tab: comparison with PI-DeepONets} summarizes the relative error for all test cases. The definition of relative error and the results of other metrics to quantify the error are provided in ~\ref{app: performance metrics}. In the upcoming sections, we take a deep dive into each test case. The code and data for all test cases are available on Zenodo \cite{b_sumanth_kumar_2024_14330861} and GitHub\footnote{\texttt{https://github.com/quest-lab-iisc/PINTO}}. 
\begin{table}[!ht]
    \centering
    \caption{\textbf{Performance Metrics:} Mean relative error (with standard deviation) of the trained PINTO models' solution when compared with the ground truth (numerical or analytical solution) for seen and unseen initial/boundary conditions. The metrics for PI-DeepONet is shown for comparison.}
    \label{tab: comparison with PI-DeepONets}
    \begin{tabular}{|c|c|c|c|c|}
    \hline
        \multirow{2}{*} & \multicolumn{2}{|c|}{\textbf{PINTO}} &  \multicolumn{2}{|c|}{\textbf{PI-DeepONet}}\\
        \hline
        \textbf{Test Case}& \textbf{Seen Conditions} & \textbf{Unseen Conditions} & \textbf{Seen Condtions} & \textbf{Unseen Conditions}\\
        \hline
        $1$D Advection equation & $2.11\%\,\left(4.01\%\right)$ & $2.85\%\,\left(4.73\%\right)$ & $1.35\%\,\left(3.75\%\right)$ & $11.26\%\,\left(11.42\%\right)$ \\
        \hline
        $1$D Burgers equation & $4.81\%\,\left(4.43\%\right)$ &$5.24\%\,\left(4.51\%\right)$ &$12.81\%\,\left(15.03\%\right)$&$11.85\%\,\left(10.78\%\right)$\\
        \hline
        Kovasznay Flow & $0.037\%\,\left(0.0325\%\right)$ & $0.41\%\,\left(2.55\%\right)$ & $0.08\%\,\left(0.066\%\right)$ & $2.26\%\,\left(6.54\%\right)$\\
        \hline
        Beltrami Flow & $0.53\%\,\left(0.9\%\right)$ & $0.6\%\,\left(0.92\%\right)$ & $2.62\%\,\left(4.19\%\right)$ & $4.89\%\,\left(12.14\%\right)$ \\
        \hline
        Lid Driven Cavity Flow & $1.36\%\,\left(1.44\%\right)$ & $2.78\%\,\left(2.49\%\right)$ & $1.96\%\,\left(2.31\%\right)$ & $6.08\%\,\left(6.61\%\right)$\\
        \hline
    \end{tabular}
\end{table}


\subsection{Advection Equation}\label{sec: Advection_equation}
As the first test case, we use PINTO to solve the one-dimensional advection equation. The advection equation is a hyperbolic equation used widely for simulating the movement of a conserved scalar in a flow field, 
\begin{subequations}\label{eq: Adv_eq}
    \begin{align}
        &\frac{\partial u}{\partial t} + \beta \frac{\partial u}{\partial x} =0,\, \text{in }\Omega=\{(x, t): x\in[0, 1], t\in (0, \infty)\}\\
        &u(0, x) = \sum_{\omega_{i}=\omega_{1},\cdots,\omega_{N}} \left(a_{i}\sin{\omega_{i}x + \phi_{i}}\right),\, \text{in } \partial\Omega \label{eq: sinusoidal_superposition}
    \end{align} 
\end{subequations}
where $x$ is the spatial coordinate, $t$ is the temporal coordinate, and $\beta$ is the constant advection speed. Different initial conditions are generated using the superposition of the sinusoidal wave equation (Eq.~\ref{eq: sinusoidal_superposition}), where $\omega_{i}=2\pi\{n_{i}\}/\Delta_{x}, i=1, \cdots, N$ whose wave numbers $\{n_{i}\}$ are random integers in $[1, n_{max}]$, $N$ is an integer that determines how many waves to add, $\Delta_{x}$ is the domain size, $a_{i}$ is the amplitude randomly chosen in $[0, 1]$, and $\phi_{i}$ is the phase randomly chosen in $(0, 2\pi)$. We set $\beta=0.1$ for all experiments. We selected 100 initial conditions. For 80 of these initial conditions, the PINTO model is trained using the physics loss Eq.~\ref{eq:mod_PINNs_loss}. The remaining 20 initial conditions are the unseen situations for which we will evaluate the initial condition generalization ability of PINTO. For validation, we use the PDEBENCH dataset that contains the numerical solutions of the advection equation for these initial conditions solved using a finite difference scheme with 2nd order upwind for time marching and a spatial grid size of $1024$ \cite{takamoto2022pdebench}. Note that the PINTO model is not trained using the PDEBENCH dataset. 



\begin{figure}[!ht]
\centering
\includegraphics[width=\columnwidth]{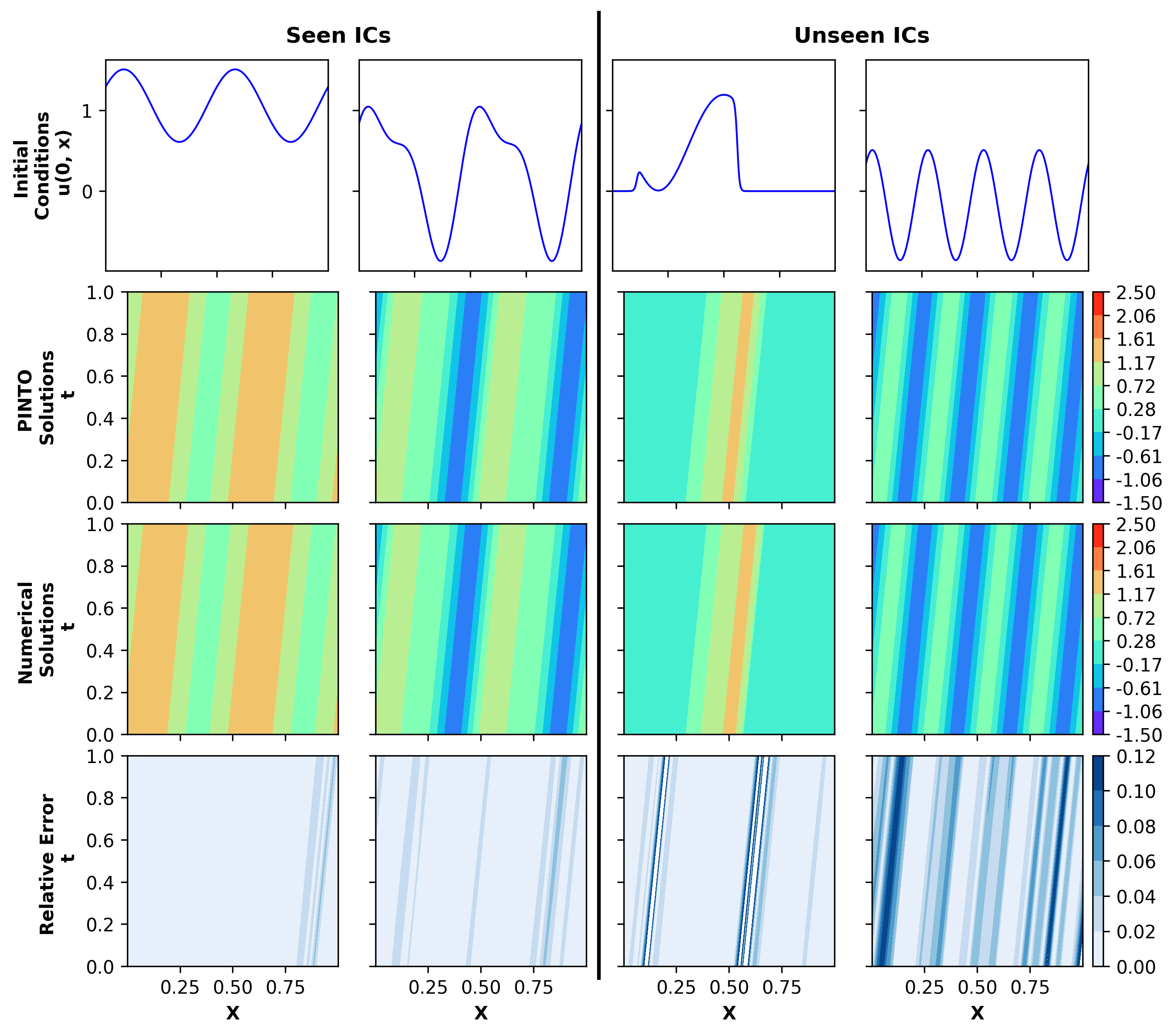} 
\caption{\textbf{Advection Equation:} Initial conditions (first row), PINTO solutions (second row), corresponding PDEBENCH data (third row) and relative error of PINTO solution (fourth row) for seen (first two columns) and unseen (last two columns) initial conditions.}
\label{fig: Advection_training_testing_plots}
\end{figure}

\begin{figure}[!ht]
\centering
\includegraphics[width=\columnwidth]{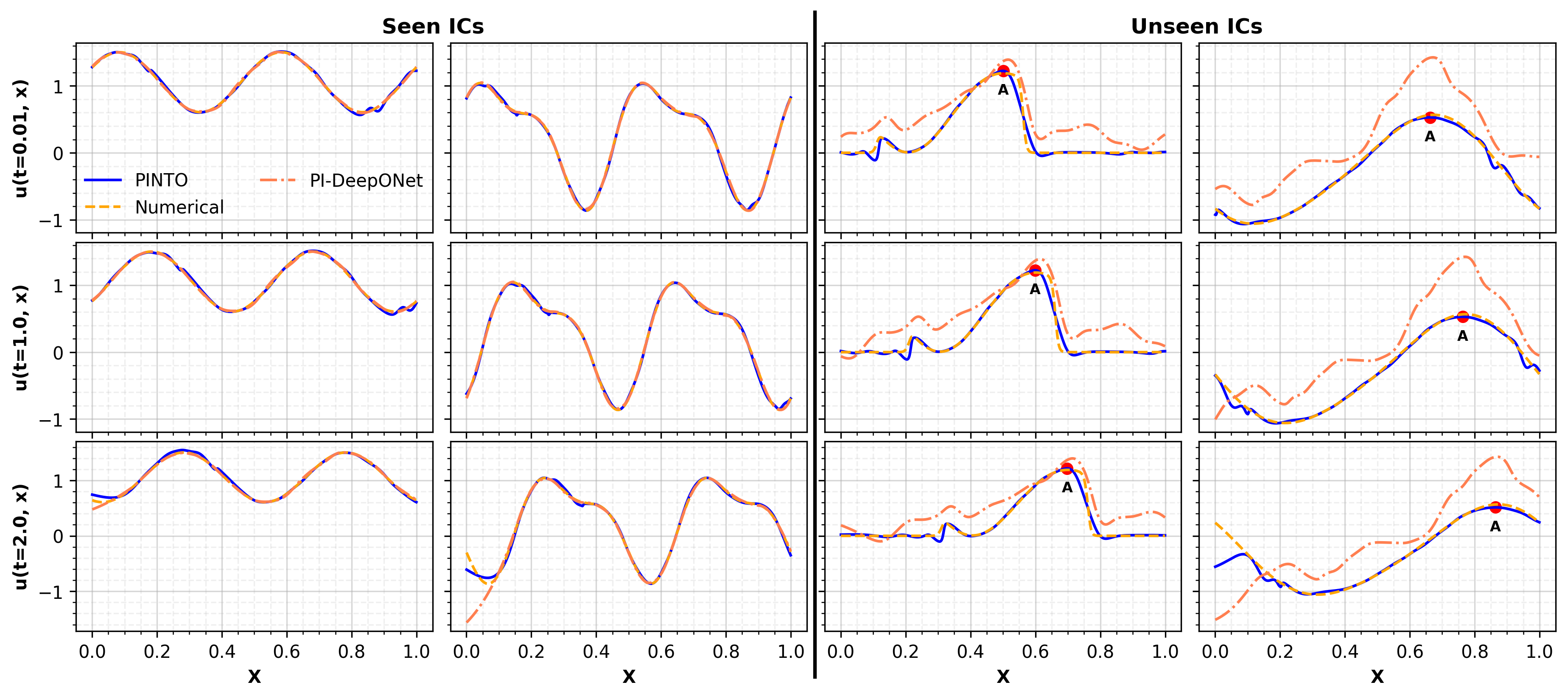} 
\caption{\textbf{Advection Equation:} PINTO, PI-DeepONet and numerical solutions (PDEBENCH) for seen and unseen initial conditions at t=$0.01,\,0.5,\,2$. The first two columns are results for seen and the last columns are for the unseen initial conditions. A landmark A is shown in the solution for unseen initial conditions (ICs) to visualize how the wave is propagating in time.}
\label{fig: Advection_temporal_slices}
\end{figure}

For training PINTO, $2000$ collocation, and $250$ initial and boundary points are considered in the domain $\left[0, 1\right]\times\left[0, 1\right]$. For comparison, the PI-DeepONet model is also trained under the same conditions. PINTO predictions, numerical solutions, and relative error across the domain for the two seen and unseen initial conditions are shown in Figure~\ref{fig: Advection_training_testing_plots}. We see that the relative error is low for both seen and unseen initial conditions. PINTO is also able to forecast the solution for times $t>1$ not included in the training time steps. Figure~\ref{fig: Advection_temporal_slices} shows the solution wave at $t=0.01,\,1.0,\,2.0$ for two seen and unseen initial conditions, superimposed with the numerical solution from PDEBENCH and the solution from PI-DeepONet.  For seen initial conditions, both models generalized well for unseen future time $(t>1)$, whereas for unseen initial conditions, only PINTO predictions are accurate for $t>1$. The relative errors for all test cases (categorized into seen and unseen initial conditions) of the PINTO solutions and the PI-DeepONet solutions for the advection equation are tabulated in Table~\ref{tab: comparison with PI-DeepONets}. The PINTO solution has a smaller relative error of $2.85\%$ compared to the PI-DeepONets $11.26\%$. 



\begin{figure}[!ht]
    \centering
    \includegraphics[width=0.9\columnwidth]{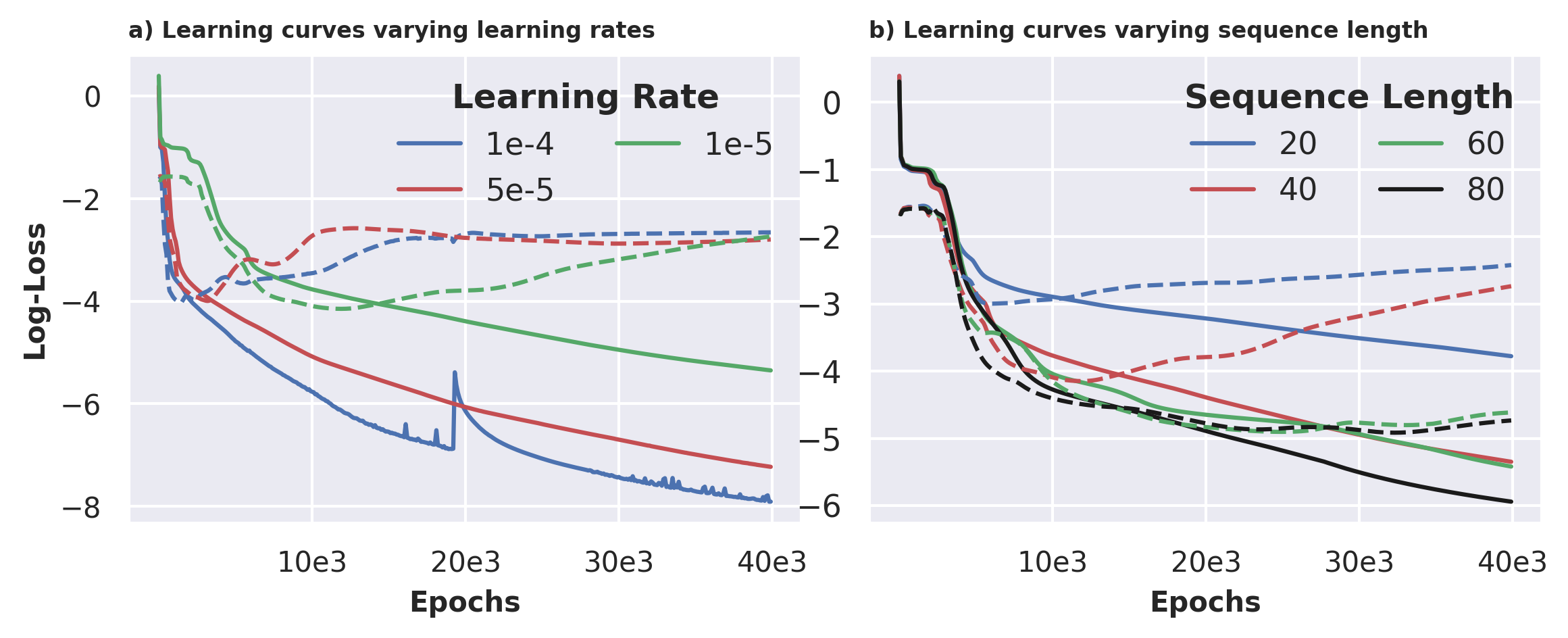}
    \caption{\textbf{Learning curves during PINTO training for advection equation:} a) Training (solid line) and validation (dashed line) loss curves for different learning rates. b) Training (solid line) and validation (dashed line) loss curves for different sequence lengths of initial conditions (input to the BPE/BVE) for a learning rate of $1e-5$.}
    \label{fig: Advection_learning_curves}
\end{figure}

Hyperparameter tuning was performed to select the number of CAUs, sequence length, learning rate, and activation functions using validation on unseen initial conditions, as described in Table~\ref{tab: adv_bur_hyp_study}. Figure~\ref{fig: Advection_learning_curves} shows the learning curves of the PINTO model for various learning rates and sequence lengths. Figure~\ref{fig: Advection_learning_curves} shows that lower learning rates and longer sequence lengths result in reduced validation loss. This indicates that longer sequence lengths are generally better for achieving generalization to unseen initial/boundary conditions. However, lower learning rates require more epochs and longer sequences increase the compute time per epoch, leading to longer training time. Taking into account both factors, the PINTO model was trained for $20000$ epochs using the Adam optimizer with a learning rate of $1e-5$ and a sequence length of $60$, achieving a mean relative error of $2.11\%$ on the initial conditions seen and $2.85\%$ on the unseen conditions. For more details on the hyperparameters used for PINTO and PI-DeepONets, see \ref{app: hyp_adv_bur}.

\subsection{Burgers Equation}\label{sec:Burgers equation}
As the second test case, we consider the one-dimensional nonlinear Burgers equation used in modeling turbulence, fluid flows, gas dynamics and traffic flow. The Burgers equation is
\begin{subequations}
    \begin{align}
        &\frac{\partial u}{\partial t} + u\frac{\partial u}{\partial x} = \nu \frac{\partial^{2}u}{\partial x^{2}},\\
        &u(0, x) = u_{0}(x), 
    \end{align}
\end{subequations}
where $u_{0}$ is the initial conditions and $\nu$ is the viscosity coefficient. 
The initial conditions $u_{0}(x)\sim\mathrm{N}\left(0,\sigma^2\right)$ are sampled from a Gaussian random field with mean zero and covariance determined by the Laplacian \cite{li2020fourier}. We use periodic boundary conditions and set $\nu=0.01$. Here, PINTO is trained to learn the mapping between multiple initial conditions and the solution of the PDE. 
PINTO is trained for $80$ different initial conditions with physics loss (Eq.~\ref{eq:mod_PINNs_loss}) considering $2000$ collocation, $250$ initial and boundary points in the domain $\left[0, 1\right]\times\left[0, 1\right]$ and tested on $20$ unseen initial conditions. Inference is performed on seen and unseen initial conditions for trained time and untrained time $t>1$. Numerical solutions for validation are obtained using an off the shelf solver \cite{li2020fourier}.
\begin{figure}[!ht]
\centering
\includegraphics[width=0.9\columnwidth]{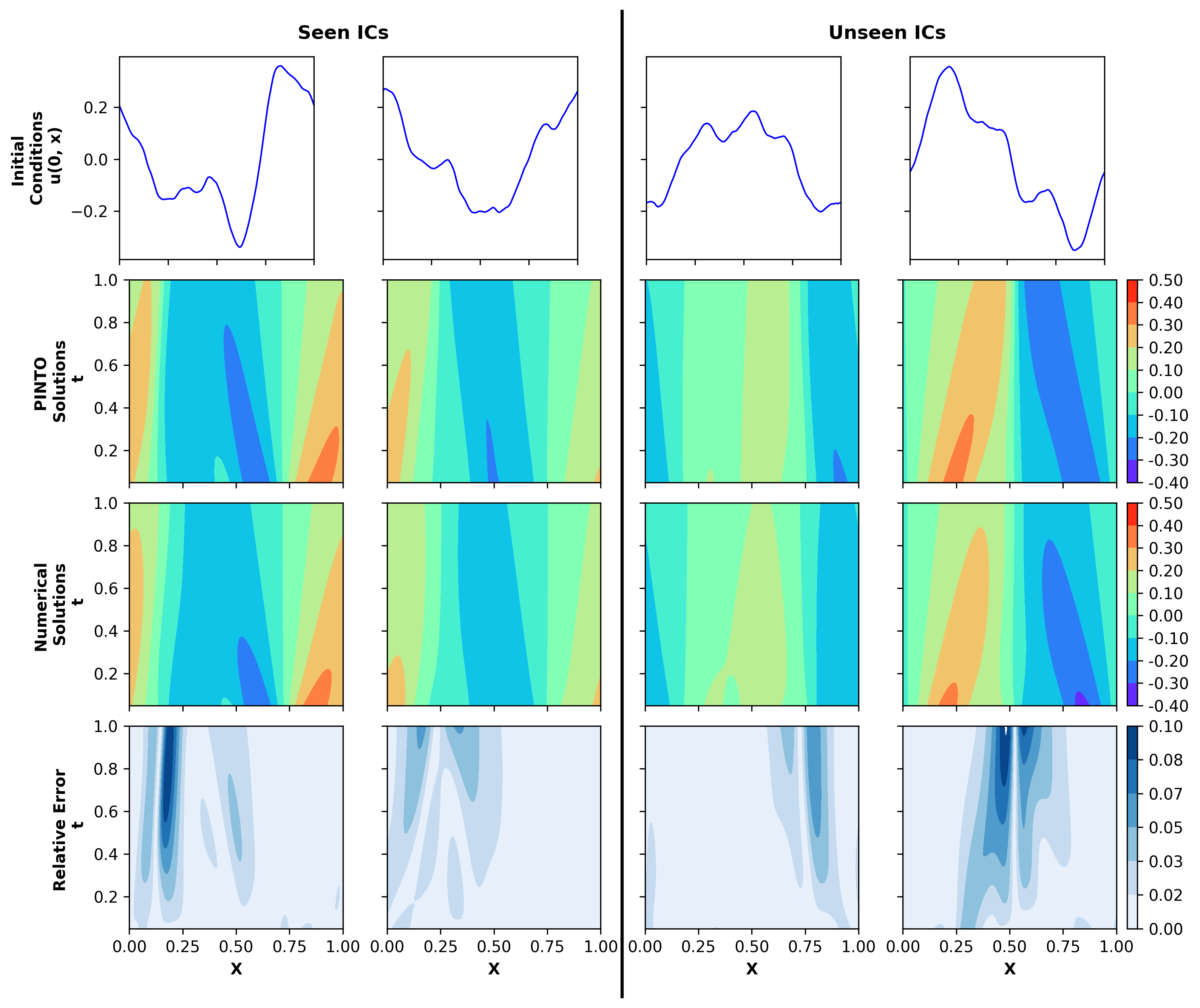} 
\caption{\textbf{Burgers Equation:} Initial conditions (first row), PINTO solutions (second row), corresponding numerical solution (third row) and relative error of PINTO solution (fourth row) for seen (first two columns) and unseen (last two columns) initial conditions.}\label{fig: Burgers_training_testing_plots}
\end{figure}

\begin{figure}[!ht]
\centering 
\includegraphics[width=\columnwidth]{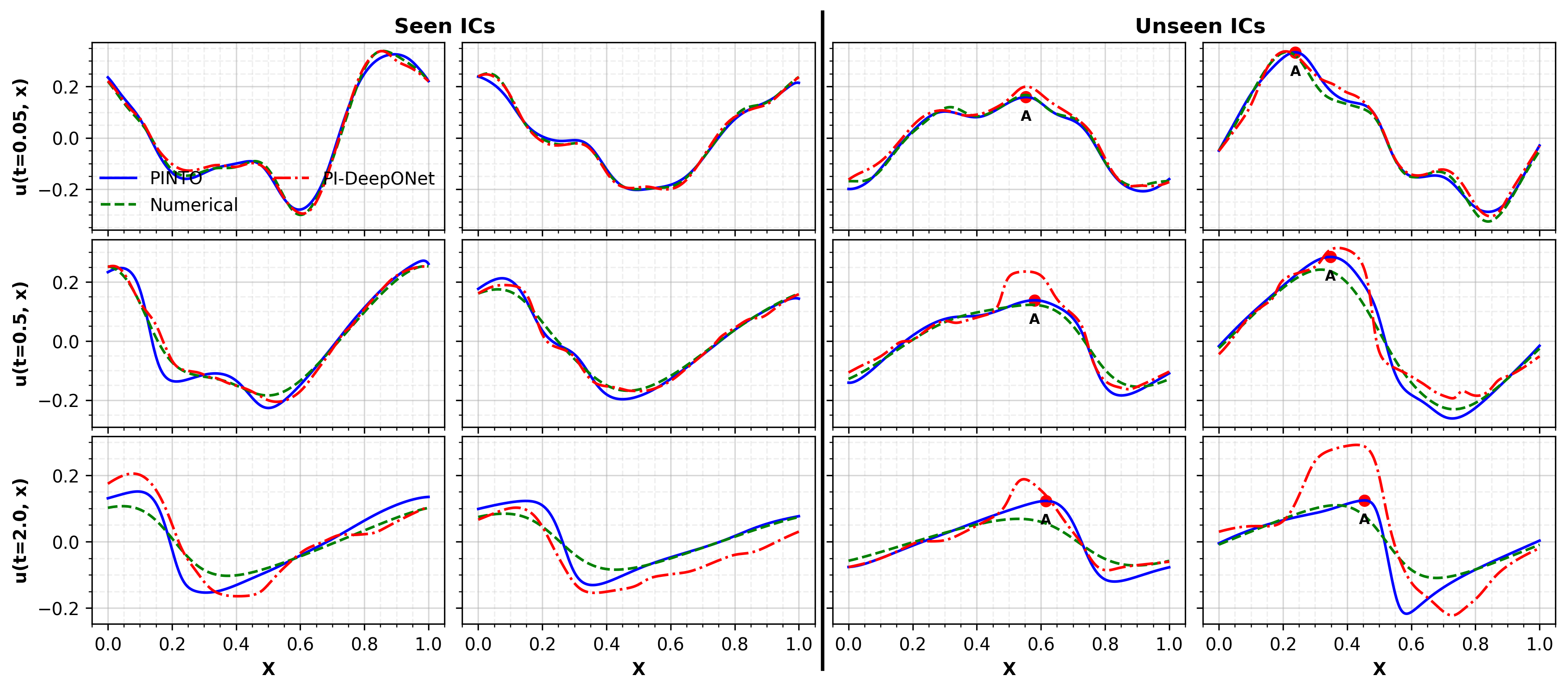} 
\caption{\textbf{Burgers Equation:} PINTO, PI-DeepONet and numerical solutions for seen and unseen initial conditions at t=$0.01,\,0.5,\,2$. The first two columns are results for seen and the last columns are for the unseen initial conditions. A landmark A is shown in the solution for unseen initial conditions (ICs) to visualize how the wave is propagating in time.}
\label{fig: Burgers_temporal_plots}
\end{figure}

Figure~\ref{fig: Burgers_training_testing_plots} shows the solution of the Burgers equation for two sets of seen and unseen initial conditions computed by PINTO. The corresponding numerical solution and the relative error are visualized to evaluate the performance of PINTO. Table~\ref{tab: comparison with PI-DeepONets} shows that the relative error for PI-DeepONet is three times that of our PINTO model for both seen and unseen initial conditions. Figure~\ref{fig: Burgers_temporal_plots} shows the solution from PINTO, numerical solver, and PI-DeepONet at three discrete times $t=0.01, 0.5, 2$ to illustrate how the solution evolves in time. $t=2$ is a time step outside the temporal domain used to train all neural models. As time increases, the deviations of the solution computed by PINTO and PI-DeepONet from the numerical solution increase, but our PINTO model has better predictions compared to PI-DeepONet and has a better generalization for unseen initial conditions. The location with maximum value is marked with the letter A in Figure~\ref{fig: Burgers_training_testing_plots} to track how the solution evolves over time. We see that PINTO is able to maintain the extreme value of the solution better than PI-DeepONet. In addition, PINTO is able to maintain periodic boundary conditions better than PI-DeepONet.

A series of experiments similar to the Advection case have been performed to determine the number of CAUs for Burgers equations. Table~\ref{tab: adv_bur_hyp_study} summarizes the experiments with different learning rates and the number of CAUs, and its last column consists of the mean relative errors in the prediction of trained models on the set of unseen initial conditions. 

\begin{figure}[!ht]
\centering
\includegraphics[width=0.8\columnwidth]{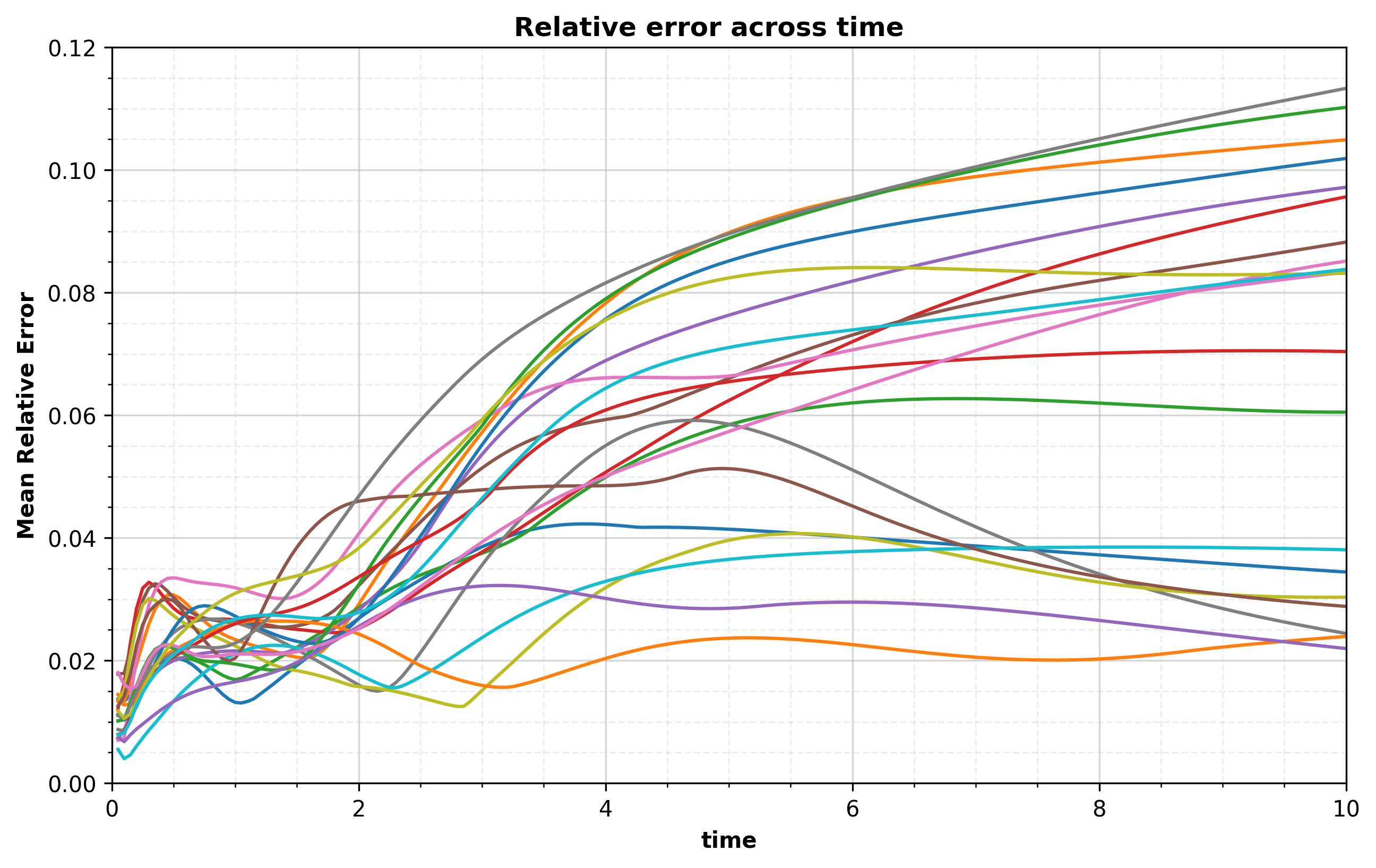} 
\caption{\textbf{Burgers Equation:} Evolution of mean relative error in time for $20$ different unseen initial conditions.}
\label{fig: burgers_error_across_time}
\end{figure}

Figure~\ref{fig: burgers_error_across_time} shows the performance of PINTO for $20$ different unseen initial conditions for longer times. Here, the mean relative error is computed for all points in the discrete spatial domain for a given time. It is observed that $2<t<6$ there is a rapid growth in the relative error, and beyond the time step $t>6$ all test cases are clustered into two groups; in one, the relative error is reaching saturation and in another, the relative error is still increasing, but with a slower rate. In all cases, the relative error is confined to 2-12\%.

\subsection{Navier-Stokes Equation}
The next three test cases, viz., Kovasznay flow, Beltrami flow and lid driven cavity flow, are solutions of the Navier-Stokes equations
 \begin{subequations}
    \begin{align}
        \frac{\partial\textbf{u}}{\partial t} + (\textbf{u}\cdot\nabla)\textbf{u} = -\nabla p + \frac{1}{Re}(\nabla^2 \textbf{u})\,,\\
        \nabla\cdot\textbf{u} = 0,\,\text{in } \Omega\,,\\
        \textbf{u} = \textbf{b},\, \text{on }\partial\Omega\,,
    \end{align}
    \label{eq: Ns}
\end{subequations}
where $\mathbf{u}$ is the velocity vector, $p$ is the pressure, Re is the Reynolds number and $\mathbf{b}$ is the initial/boundary condition. For validation, we use the analytical solution of the Kovasznay and Beltrami flows, and the numerical solution from a finite volume solver for the lid driven cavity flow. In what follows, we discuss each test case in detail.

\subsubsection{Kovasznay Flow}\label{sec: Kovasznay_flow}
\begin{figure}[!ht]
\centering
\includegraphics[width=\columnwidth]{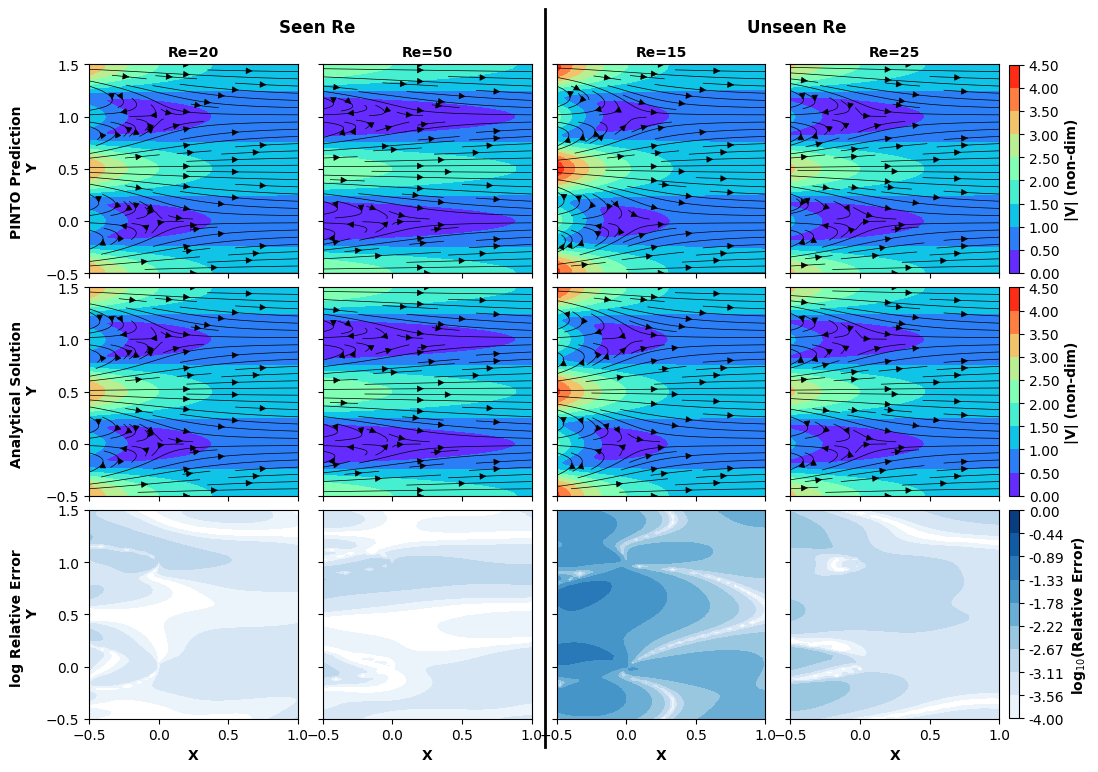} 
\caption{\textbf{Kovasznay Flow:} PINTO solutions (first row), corresponding analytical solution (second row) and relative error of PINTO solution (fourth row) for seen (first two columns) and unseen (last two columns) initial conditions (Re). Flow streamlines are overlaid on a background of the velocity magnitude $|V|=\sqrt{u^{2} + v^{2}}$ for the solution.  As the error is small, $\log_{10}$ of magnitude of the relative error is shown.}
\label{fig: kovasznay_training_plots}
\end{figure}
The Kovasznay flow is governed by the steady-state Navier-Stokes equation (first term of eq.~\ref{eq: Ns}a is 0) and has the following analytical solution
\begin{align*}\label{eq: Kovasznay_sol}
    u(x, y) &= 1 - e^{\eta x} \cos(2 \pi y), \\
    v(x, y) &= \frac{\eta}{2 \pi} e^{\eta x} \sin(2 \pi y), \\
    p(x, y) &= \frac{1}{2} (1 - e^{2 \eta x})\,,
\end{align*}
where $\eta=0.5\text{Re}-\sqrt{0.25\text{Re}^2+4\pi^2}$. We consider the domain $\left[-0.5, 1.0\right] \times \left[-0.5, 1.5\right]$. To numerically solve the Navier-Stokes equations for the Kovasznay flow, boundary conditions are provided to the solver by evaluating the analytical solution at the boundary. Changing the value of Re changes the boundary condition. In this test case, PINTO learns the mapping between the boundary conditions and the solution of the PDE.

For training PINTO, $2000$ domain collocation points and $254$ boundary points are used. A sequence length of $80$ is used for the BPE and BVE units. The boundary values of the BVE input sequence are $x$-directional velocity, $y$-directional velocity, and pressure. As the output is the vector ($u, v, p$), PINTO has three separate output units for $u$, $v$, and $p$, each with $2$ hidden dense layers ($64$ neurons). Training is performed by minimizing the physics loss Eq.~\ref{eq:mod_PINNs_loss} for $40,000$ epochs using the Adam optimizer with a learning rate of $5e-4$. The PI-DeepONet is also trained in the same setting for comparison. We used the relative error in the total velocity magnitude $\left(|V| = \sqrt{u^{2} + v^{2}}\right)$ as a performance metric to evaluate PINTO and PI-DeepONet.

To examine the generalization of the neural model to various boundary conditions in the Kovasznay flow, both the PINTO and PI-DeepONet models are first trained for boundary conditions corresponding to Re = 20, 30, 50, and 80. The trained model is then used to simulate the solution for unseen boundary conditions corresponding to 20 different Reynold's numbers, which are randomly generated between 10 and 100. Table~\ref{tab: comparison with PI-DeepONets} presents the mean relative error comparisons between the trained PINTO and PI-DeepONet model simulations for both seen and unseen boundary conditions. Overall, the relative error is low for both neural architectures for the seen boundary conditions. The relative error for PI-DeepONet is 2 times for seen and 5 times for unseen conditions, compared to our PINTO model, showing the latter's superior performance.

The 2D field of the solution is visualized using a $64 \times 64$ grid (Figure~\ref{fig: kovasznay_training_plots}). The PINTO solution, the analytical solution, and the relative error are shown for two sets of seen (Re = $20$ and $50$) and unseen (Re = $15$ and $25$) boundary conditions. As the Kovasznay flow is a steady flow, time is not considered here. We see that the visual quality of the solution matches the low relative error for our PINTO model reported in Table~\ref{tab: comparison with PI-DeepONets}. Overall, the findings demonstrate that our PINTO model can successfully generalize to unseen boundary conditions.

\subsubsection{Beltrami Flow}\label{sec:betrami_flow}
\begin{figure}[!ht]
\centering
\includegraphics[width=0.9\columnwidth]{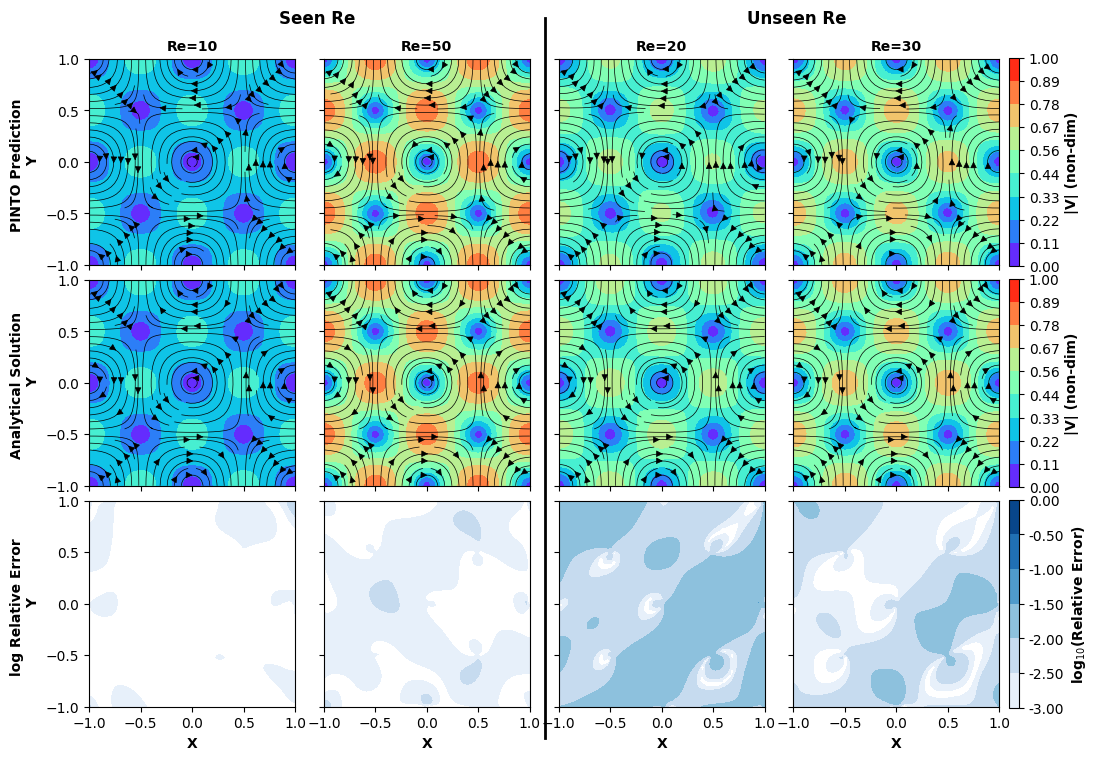} 
\caption{\textbf{Beltrami Flow:} PINTO solutions (first row), corresponding analytical solution (second row) and relative error of PINTO solution (fourth row) for seen (first two columns) and unseen (last two columns) initial conditions (Re) at solution time step $t=0.5$. Flow streamlines are overlaid on a background of the velocity magnitude $|V|=\sqrt{u^{2} + v^{2}}$ for the solution. As the error is small, $\log_{10}$ of magnitude of the relative error is shown so that it is visible.}
\label{fig: Beltrami_training_plots}
\end{figure}

The Beltrami flow is governed by the unsteady Navier-Stokes equation (Eq.~\ref{eq: Ns}), with a dynamic spatially varying boundary condition. This flow has an analytical solution given by
\begin{align*}
    u(x, y, t) &= -\cos{\pi x}\,\sin{\pi y}\,e^{-2 \pi^{2}\nu t},\\
    v(x, y, t) &= sin{\pi x}\,\cos{\pi y}\,e^{-2 \pi^{2}\nu t},\\
    p(x, y, t) &= -\frac{\cos{2 \pi x} + \cos{2 \pi y}}{4}\,e^{-4 \pi^{2}\nu t}, 
\end{align*}
where $\nu = 1/\text{Re}$. Similar to the previous test case, the numerical solvers are given the initial and boundary condition obtained from the analytical expression. Changing the Reynolds number gives multiple sets of initial and boundary conditions. 
In total, $5000$ collocation points are sampled from the given spatiotemporal domain, $1000$ boundary condition points ($250$ from each of the 4 sides), and $500$ initial condition points. The same model configuration and hyperparameter choice as for the PINTO model for Kovasznay flow is used here. 

The PINTO model was trained for boundary conditions corresponding to Re = 10, 50, and 100. Subsequently, simulations are performed using the trained model for unseen boundary conditions corresponding to $20$ different Re generated randomly between 10 and 150. In Figure~\ref{fig: Beltrami_training_plots}, the first two columns show the comparison between the predictions from PINTO and the analytical solution for seen boundary conditions during training, and the last two columns show the comparison for unseen boundary conditions during testing. Table~\ref{tab: comparison with PI-DeepONets} summarizes the mean relative error in the time interval of $[0, 2]$ for the seen and unseen boundary conditions.

\begin{figure}[!ht]
\centering
\includegraphics[width=0.8\columnwidth]{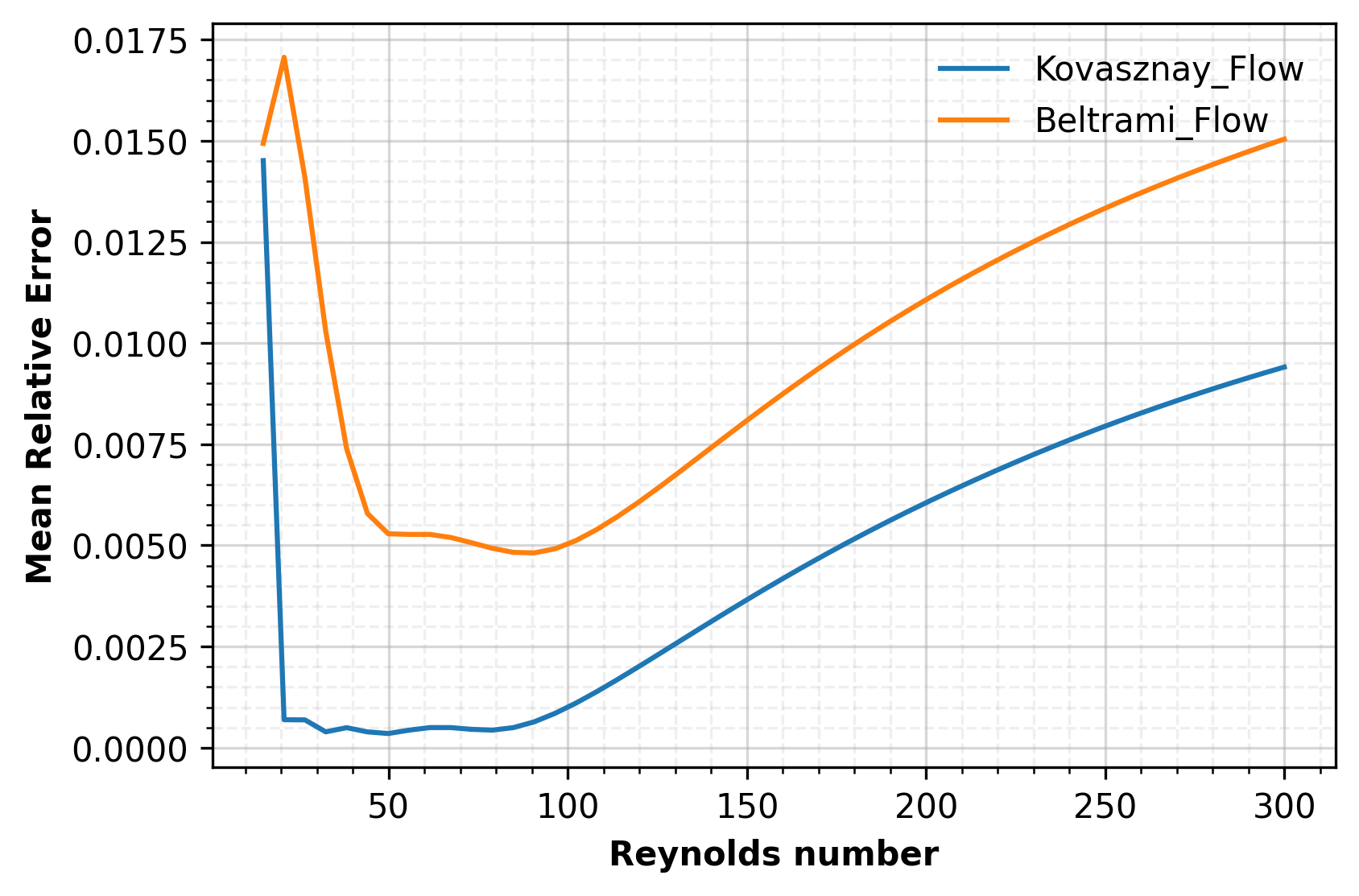} 
\caption{Mean relative error of PINTO predictions for the Kovasznay and Beltrami flows at different Reynolds numbers.}
\label{fig: relative_across_Re}
\end{figure}
Figure~\ref{fig: relative_across_Re} shows the performance of PINTO at different Reynolds numbers (Re). For the Kovasznay flow, the mean relative error for a given Re is computed across all discrete points in the spatial domain. For the Beltrami flow, for a given Re, the mean relative error is computed across all discrete spatio-temporal points in the spatio-temporal domain. The minimum and maximum Re considered to train the PINTO model are $20,\,80$ for the Kovasznay and $10,\,100$ for the Beltrami Flow. In both cases, as Re increases, there is an increase in the mean relative error, but it is less than 1.5\% even at very large unseen Re that were not shown during the training. These results demonstrate that PINTO training is robust and generalizable.
\subsubsection{Lid Driven Cavity Flow} \label{sec:lid_driven_flow}
\begin{figure}[!ht]
\centering
\includegraphics[width=0.9\columnwidth]{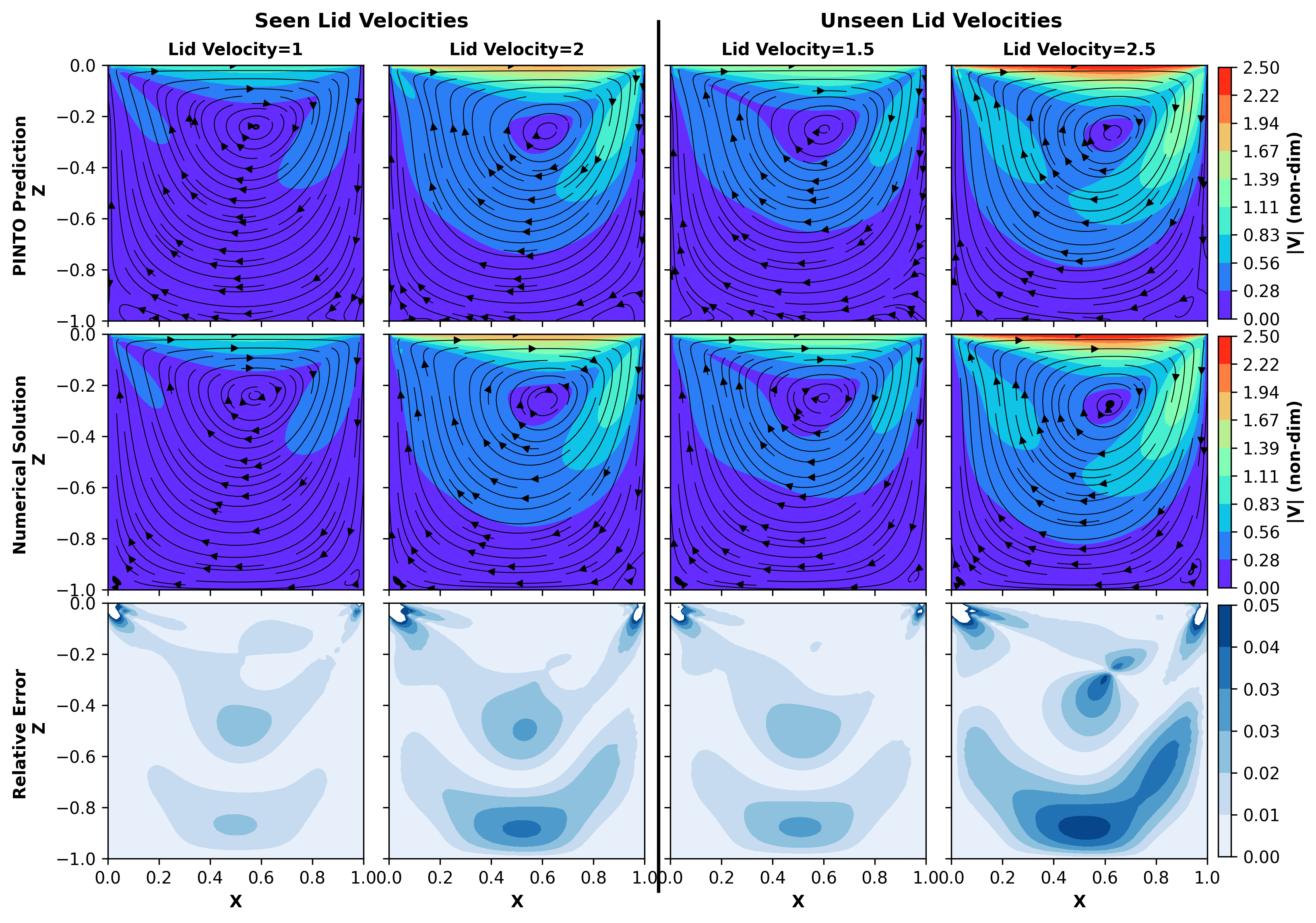} 
\caption{\textbf{Lid Driven Cavity Flow:} PINTO solutions (first row), corresponding analytical solution (second row) and relative error of PINTO solution (fourth row) for seen (first two columns) and unseen (last two columns) boundary conditions (Lid Velocity). Flow streamlines are overlaid on a background of the velocity magnitude $|V|=\sqrt{u^{2} + v^{2}}$ for the solution. Only magnitude of relative error is shown.}
\label{fig: Lid_driven_plots}
\end{figure}
We simulate the steady-state Navier-Stokes equation for the lid-driven flow at a Reynolds number of $50$ (Eq.~\ref{eq: Ns} with first term zero, Re=50) in a computational domain $\left[0, 1\right] \times \left[0, 1\right]$. Since the only boundary that changes is the lid's velocity, here we learn the mapping between lid velocities (the top boundary condition) to the solution space. In the domain, we considered $2000$ collocation points and $400$ boundary points ($100$ on each side of the boundary). Here, the boundary condition is the velocity of the lid. During training, we used the lid velocities of 1, 2, and 3. Figure~\ref{fig: Lid_driven_plots} shows the comparison plots between PINTO prediction and numerical solutions for seen lid velocities of 1 and 2 during training in the first two columns and for unseen lid velocities of 1.5 and 2.5 in the last two columns inside. 
\begin{figure}[!ht]
\centering
\includegraphics[width=0.9\columnwidth]{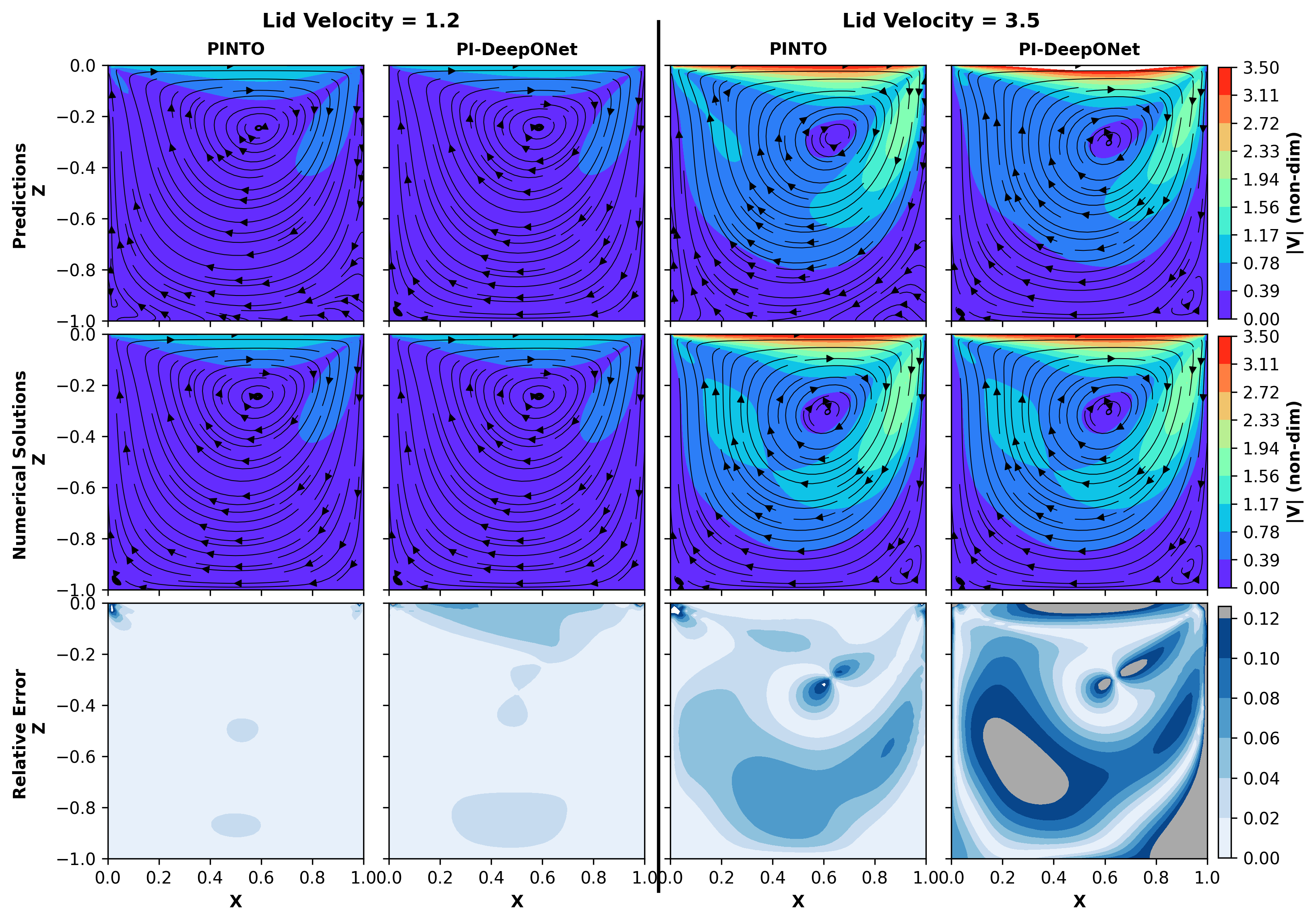} 
\caption{\textbf{Lid Driven Cavity Flow:} PINTO and PI-DeepONet solutions (first row), corresponding numerical solution (second row) and relative error (third row) for unseen lid velocity interpolation (1.2) and extrapolation (3.5). Flow streamlines are overlaid on a background of the velocity magnitude $|V|=\sqrt{u^{2} + v^{2}}$ for the solution. Only magnitude of relative error is shown.}
\label{fig: Lid_driven_comp_plots}
\end{figure}
Table~\ref{tab: comparison with PI-DeepONets} documents the mean relative error in the predictions of the PINTO on the $64 \times 64$ grid for the seen and unseen lid velocities. For validation, numerical solutions for all lid velocities during training and testing are simulated from the Finite Volume code \citep{ueckermann_and_lermusiaux_MSEAS2012}.
Figure ~\ref{fig: Lid_driven_comp_plots} compares PINTO and PI-DeepONet solutions for unseen lid velocities between 1 and 3 (the values seen during training) and greater than 3 (outside the range of values seen during training). For the lid velocity of $1.2$, the PINTO and PI-DeepONet solutions are accurate and comparable to the numerical solution, whereas, for the lid velocity of $3.5$, the PINTO solutions are better compared to the PI-DeepONet solutions. For PI-DeepONet, in this challenging test, the relative error crosses 12\% at several grid points, making the solution unusable. In general, the relative error for different unseen lid velocities from the PINTO solution is $2.78\%$, which is less than half of the relative error ($6.08\%$) of the PI-DeepONet solution.  

\section{Discussion} \label{sec:discussion}
In this section, we discuss computational complexity, the need for a neural operator, the interpretation of our neural operator from the point of view of classical numerics, comparison to other neural operators, insights into how the cross-attention unit can enhance other neural operators, and limitations.

\textbf{Computational complexity:} Generally, transformer operators are computationally intensive due to the attention matrix. In addition, transformer-based neural operators \cite{hao2023gnot} require extensive simulation or observational data to learn to generalize to irregular meshes and multiple input functions. In contrast, PINTO achieves generalization using cross-attention units trained by minimizing the physics-loss. As the query point is a sequence of length $M=1$, the complexity of PINTO's attention mechanism is $O\left(N\times m^{2}\right)$, as opposed to $O\left(MN\times m^{2}\right)$ of general attention, where $M, N$ represents query and key sequence lengths and $m$ is the embedding dimension (dimension of encoded vector). \ref{app: computational-time} shows the computational time for all test cases.


\textbf{Need of a neural operator:}
Vanilla PINNs learn a mapping between the input domain and the solution space $\mathcal{G}:\Omega\mapsto\mathcal{H}$ that satisfies the differential operator for a specific initial/boundary condition. Here, $\mathcal{G}_\theta(\mathbf{x};\Theta)$ is a deep neural network with $\Theta$ as parameters that approximate $h\in\mathcal{H}$ of Eq.~\ref{eq:def:pde}. To obtain the optimal parameters of the network $(\Theta^{*})$, we solve the following optimization problem. 
\begin{equation}
    \min_{\theta}\,\frac{1}{N_{c}}\sum_{j=1}^{N_{c}}|f^{j}-\mathcal{N}(\mathcal{G}(\Theta;X^{j});\alpha)|^{2} + \frac{1}{N_{ib}}\sum_{j=1}^{N_{ib}}|b^{j}-\mathcal{B}(\mathcal{G}(\Theta;X_{b}^{j}))|^{2}\,,
\end{equation}\label{eq:func:PINNs_loss}
where, $\{(X_{b}^{j}, b^{j})\}_{j=1}^{N_{ib}}$ denote the training data points on $\partial\Omega$, and $\{X^{j}\}_{j=1}^{N_{c}}$ are the training points (called collocation points) on the domain $\Omega$. $\mathcal{G}(X;\Theta^{*})$ is an approximate solution to Eq.~\ref{eq:def:pde} for a given $b$. For a different $b$, the optimization problem in Eq.~\ref{eq:func:PINNs_loss} must be solved again to obtain another set of optimal parameters $\left(\Theta^*\right)$. In other words, PINNs need retraining if the initial and boundary conditions change. A neural operator framework is necessary to generalize to different initial and boundary conditions. However, the neural operators in the literature are majorly trained using simulation data and recently with some works reporting physics-informed training. In the present work, our focus is on developing a neural operator that efficiently generalizes to different initial/boundary conditions while training only with physics loss. 

\textbf{Interpretation from classical numerics:}
Neural operators are akin to traditional numerical techniques, such as finite elements, substituting the linear span of local basis functions with neural operator layers \cite{li2020fourier}. The initial and boundary conditions ensure the necessary constraints for a unique PDE solution, dictating the function space for the solution, so the network must learn to incorporate these conditions. Numerical methods involve discretizing the PDE Eq.~\ref{eq:def:pde} into a system of linear equations $(Ax = b)$, providing solutions across the grid. Alternatively, the point solution corresponds to the weighted sum of the initial and boundary conditions ($x = A^{-1}b$). Interior points as queries, with initial and boundary points as keys and values in an attention mechanism, yield the weighted sum of boundary functions, mirroring numerical methods. The attention score combines two terms: \textit{(i)} similarity and \textit{(ii)} magnitude \cite{song2021implicit}. The similarity term uses an RBF kernel with a length-scale hyperparameter to assess the query-key importance. Thus, our PINTO model can be viewed as a learning scheme inspired by traditional numerical methods, to use the attention mechanism for learning boundary-aware vectors. For the advection equation we show an explicit calculation in ~\ref{app: interpretation_FDM} to illustrate the above interpretation.

\textbf{Comparison to other neural operators:} 
It is worth noting that traditional physics-informed neural networks require separate training for each set of initial/boundary conditions, whereas our PINTO model allows training once for multiple initial/boundary conditions and can handle unseen conditions during inference. The physics-informed DeepONet (PI-DeepONet) \citep{lu2021learning,goswami2023physics} used in the present paper for the comparison of PINTO was repurposed to find PDE solutions with different initial/boundary conditions. 
In DeepONets, the vector of input functions is passed through the branch network. The branch network is an MLP, with the input layer dimension equal to the length of the vector; since the shape of the weight matrix is fixed between the input layer and the first hidden dense layer, it is not possible to pass vectors of different lengths to the DeepONets during the inference phase. On the other hand, the attention mechanism processes the input as a sequence of tokens. Since PINTO uses an attention mechanism, it allows different vector lengths as input for the BPE and BVE units during training and testing rather than a fixed length without affecting the model architecture. By means of the cross-attention mechanism, the query point's representation vector effectively captures all the relevant information from the boundary points and their values, which is necessary for predicting the PDE solution at that specific query point for any unseen boundary condition. As our goal was to train with physics-loss, other neural operator models such as physics-informed neural operators, Fourier neural operators, or convolutional neural operators that need simulation data for training at some resolution are not considered for comparison. 

Unlike the DeepONet architecture \citep{lu2021learning, wang2021learning}, one key novelty of PINTO is the treatment of the initial/boundary conditions as a sequence of vectors, where each element is a vector of initial/boundary coordinates and values. In addition, PINTO uses an attention mechanism to learn the encoded representation of the initial/boundary functional space $\mathcal{B}$. This mode of representation not only helps in generalization but also gives the practical utility of using different lengths of boundary vectors during training and inference phases, making our approach further different from the branch and trunk nets (DeepONet). 

\textbf{Cross Attention Units in other neural operators:} The kernel integral operator implemented using the cross-attention unit of PINTO is versatile. In PINTO, we used attention in the representation space of the domain's coordinates and the initial/boundary conditions. In the Fourier Neural Operator, this mechanism can be incorporated by performing a cross attention between the input field and solution domain in the Fourier space. Time marching neural operators such as OFormer and GNOs can also benefit from the introduction of cross-attention units to condition the representation space with input functions.

\textbf{Extending generalization to other parameters:} In this paper we emphasized generalization for any initial and boundary conditions. However, the idea of the cross attention unit can be extended to other types of generalization such as geometry or other flow parameters. For example, if the position of obstacles or shape of objects in a flow field can be parametrized to a functional representation, then PINTO could be applied to the parametrized sequence of geometry. For generalizing for unseen fluid properties (such as viscosity), the cross attention units can attend to the sequence of parametrized properties. 

\textbf{Extension to 3D in space test cases:} The extension of PINTO to 3D in space is straightforward. The number of collocation points and boundary points will increase, leading to an increase in the training time and memory requirement. An increase in training instabilities due to vanishing or exploding gradients, and imbalances in the loss terms, could also arise. Furthermore, the complexity of the cross-attention units is more than PI-DeepONet, thus training will be more expensive in comparison.

\textbf{Physics guided training and discrete loss functions:} In this paper, we focus on physics-informed training using automatic differentiation. However, the proposed architecture can also be trained using simulation data in a physics-guided setting. ~\ref{app: data-driven-training} shows results for the physics-guided training of Burger's equation. Instead of automatic differentiation, discrete loss functions based on finite difference, finite volume \citep{rishi2024training, li2025learning} and finite element \citep{anandh2024efficient, yamazaki2025finite}

\textbf{Limitations:} Since the training of the PINTO model uses collocation points and physics-loss, it may suffer spectral bias, a common phenomenon while training neural networks, and has also been reported in other PINN models. There might also be training discrepancies due to imbalances in different terms in the loss functions. Adaptive training, adaptive sampling, and dynamic weighting strategies \citep{jin2021nsfnets, 10.5555/3618408.3618696, hou2023enhancing, maddu2022inverse, deguchi2023dynamic, tang2023pinns, moseley2023finite, wang2021understanding} must be adopted to address these limitations.

\section{Conclusion and Future Scope} \label{sec:conclusion}
Our newly developed PINTO architecture enhances the generalizability of neural models to solve PDEs with multiple initial and boundary conditions. The key novelty of the architecture is the introduction of the new cross-attention unit that implements the attention-based integral kernel operator. This architecture is successful in transforming the representation vector of the PDE solution's domain to a boundary/initial condition-aware encoding. Using five diverse test cases, we demonstrated that the relative error of the PINTO model is comparable for seen and unseen initial and boundary conditions. These results show that the encoding learned through our cross-attention unit is efficient in generalizing to new out-of-distribution (unseen during training) input functions of initial and boundary conditions. Comparative analysis with the PI-DeepONet neural operator model demonstrates the superior performance of PINTO to effectively generalize to unseen initial and boundary conditions. The relative error of PINTO solutions for all test cases in unseen testing configurations is low, only 20\% to 33\% of the relative error of PI-DeepONet's solutions. We also show that PINTO is able to find solutions outside the temporal domain in which training is performed, the first result of this nature in the neural operator literature. Future directions of our work include the incorporation of turbulent dynamics into the PINTO framework. We envision that the PINTO model will find applications in optimizing the layout of wind farms, building digital twins, modeling the earth system, and any other domain governed by initial boundary value problems. Moreover, the cross-attention units are versatile, hence, they could be used to devise architectures for geometry generalization as well.

\section*{Acknowledgements}
We thank the members of our QUEST Lab at the Indian Institute of Science Bangalore for useful discussions, especially Rishi Jinka and Azhar Gafoor. This work was partially funded by research grants from the Ministry of Earth Sciences (MoES/36/OOIS/Extra/84/2022). 

\appendix
\section{Performance metrics}\label{app: performance metrics}
In this appendix, we define the relative error metric shown in Section \ref{sec:applications} and other error metrics, and report results to analyze the PINTO and PI-DeepONet models. Section \ref{sec:applications} presents results exclusively for the adjusted relative error. This adjustment (Eq.~\ref{eq:rel_err}) is implemented to handle scenarios in which true values are nearly zero; in these instances, the absolute error is employed. This approach enhances the interpretability of the model's performance.  The definitions of all error metrics used in the paper are as follows.
{\allowdisplaybreaks
\begin{align}\label{eq: Performance metrics}
    & \text{Root Mean Squared Error(RMSE)} = \sqrt{\frac{1}{N_{q}}\sum_{i=1}^{N_{q}}(h_{i}-\hat{h}_{i})^{2}}\\
    & \text{Mean Absolute Error(MAE)} = \frac{1}{N_{q}}\sum_{i=1}^{i=N_{q}}|h_{i}-\hat{h}_{i}|\\
    & \text{Normalized Root Mean Squared Error(nRMSE)} = \frac{\text{RMSE}}{\sqrt{\frac{1}{N_{q}}\sum_{i=1}^{N_{q}}h_{i}^{2}}}\\
    & \text{Mean Absolute Percentage Error(MAPE)} = \frac{\text{MAE}}{\frac{1}{N_{q}}\sum_{i=1}^{N_{q}}|h_{i}|}\\
    & \text{Relative Error (modified)} = \frac{|h_{i}-\hat{h}_{i}|}{1+|h_{i}|} \label{eq:rel_err}\\
    & \text{Mean Relative Error (modified)} = \frac{1}{N_{q}}\sum_{i=1}^{N_{q}}\frac{|h_{i}-\hat{h}_{i}|}{1+|h_{i}|}
\end{align}}
where, $N_{q}$ is total number of query points, $\hat{h}_{i}$ is the prediction of $i^{th}$ query point from the model and $h_{i}$ is the true value of $i^{th}$ query point. 

Tables~\ref{tab: performance_metrics_1} and ~\ref{tab: performance_metrics_2} highlight that the generalization for the PINTO model is better than the PI-DeepONets across multiple performance metrics.
\begin{table}[!ht]
    \centering
    \caption{Different performance metrics for all test cases}
    \label{tab: performance_metrics_1}
    \begin{tabular}{|c|c|c|c|c|c|c|c|c|}
        \hline
        \textbf{Test Cases} & \multicolumn{4}{|c|}{\textbf{PINTO}} & \multicolumn{4}{|c|}{\textbf{PI-DeepONets}}  \\
        \hline
        & \multicolumn{2}{|c|}{\textbf{Training}} & \multicolumn{2}{|c|}{\textbf{Testing}} & \multicolumn{2}{|c|}{\textbf{Training}} & \multicolumn{2}{|c|}{\textbf{Testing}}\\
        \hline
        & \textbf{RMSE} & \textbf{MAE} & \textbf{RMSE} & \textbf{MAE} & \textbf{RMSE} & \textbf{MAE} & \textbf{RMSE} & \textbf{MAE}\\
        \hline
        \textbf{Advection} & $0.042$ & $0.026$ & $0.074$ & $0.0398$ & $0.0235$ & $0.01417$ & $0.2185$ & $0.1589$\\
        \hline
        \textbf{Burgers} & $0.0469$ & $0.0245$ & $0.0421$ & $0.0258$ & $0.0692$ & $0.0334$ & $0.0834$ & $0.0546$\\
        \hline
        \textbf{Kovasznay} & $0.0009$ & $0.0007$ & $0.00135$ & $0.001$ & $0.0022$ & $0.0016$ & $0.0435$ & $0.0269$\\
        \hline
        \textbf{Beltrami} & $0.0138$ & $0.0068$ & $0.0154$ & $0.0086$ & $0.0598$ & $0.0338$ & $0.1524$ & $0.0644$\\
        \hline
        \textbf{Lid Driven} & $0.0307$ & $0.0193$ & $0.0642$ & $0.045$ & $0.0514$ & $0.0288$ & $0.1983$ & $0.1055$\\
        \hline
    \end{tabular}
\end{table}

\begin{table}[!ht]
    \centering
    \caption{Different performance metrics for all test cases}
    \label{tab: performance_metrics_2}
    \begin{tabular}{|c|c|c|c|c|c|c|c|c|}
        \hline
        \textbf{Test Cases} & \multicolumn{4}{|c|}{\textbf{PINTO}} & \multicolumn{4}{|c|}{\textbf{PI-DeepONets}} \\
        \hline
        & \multicolumn{2}{|c|}{\textbf{Training}} & \multicolumn{2}{|c|}{\textbf{Testing}} & \multicolumn{2}{|c|}{\textbf{Training}} & \multicolumn{2}{|c|}{\textbf{Testing}}\\
        \hline
        & \textbf{nRMSE} & \textbf{MAPE} & \textbf{nRMSE} & \textbf{MAPE} & \textbf{nRMSE} & \textbf{MAPE} & \textbf{nRMSE} & \textbf{MAPE}\\
        \hline
        \textbf{Advection} & $0.0585$ & $0.0471$ & $0.1037$ & $0.0733$ & $0.0322$ & $0.0251$ & $0.3048$ & $0.2926$\\
        \hline
        \textbf{Burgers} & $0.2738$ & $0.1895$ & $0.2711$ & $0.2112$ & $0.4044$ & $0.2578$ & $0.5376$ & $0.4467$\\
        \hline
        \textbf{Kovasznay} & $0.000739$ & $0.000662$ & $0.000715$ & $0.000655$ & $0.00179$ & $0.00151$ & $0.001758$ & $0.00149$\\
        \hline
        \textbf{Beltrami} & $0.0295$ & $0.0171$ & $0.0249$ & $0.0130$ & $0.1276$ & $0.0848$ & $0.1046$ & $0.065$\\
        \hline
        \textbf{Lid Driven} & $0.0522$ & $0.0494$ & $0.0817$ & $0.073$ & $0.0874$ & $0.0739$ & $0.2523$ & $0.1708$\\
        \hline
    \end{tabular}
\end{table}
\section{Choice of hyperparameters}\label{app: hyp_adv_bur}
Like any neural model, the performance of PINTO is influenced by the number of layers, units, the activation function, and other hyperparameters. Here, the decision of the number of cross-attention units (CAUs) in the second stage of PINTO, the sequence length of initial and boundary conditions, and the choice of training hyperparameters are made based on cross validation. Since here we are emphasizing the applicability of our proposed architecture to generalize for unseen initial and boundary conditions, we have chosen the hyperparameters that gave good predictions with low relative errors on the unseen conditions; alternatively, one can choose the optimal set of hyperparameters using the grid search algorithm or efficient search algorithms \citep{bergstra2011algorithms,escapil2023hyper,kaplarevic2023identifying,wang2024pinn}. Table~\ref{tab: adv_bur_hyp_study} summarizes the hyperparameters selected for Advection and Burgers equations. For solving Navier Stokes equation, PINTO has learned the mapping with lesser relative errors (Table.~\ref{tab: comparison with PI-DeepONets}) with a single CAU unit and shallow multi-layer perceptrons for lifting and projection units with activation function of \texttt{Swish}. Due to high imbalances in the boundary values for lid-driven cavity flow, while training both PINTO and PI-DeepONet, the loss term with $0$ boundary value has given a weight of $100$.
\begin{table}[!ht]
    \centering
    \caption{Mean relative error of PINTO predictions for different hyperparameters for Advection and Burgers equation. Here, the mean of relative error across all the test cases is indicated as the validation metrics}
    \label{tab: adv_bur_hyp_study}
    \begin{tabular}{| p{0.8cm} | p{1.5cm} | p{1cm} | p{1.5cm} | p{3.5cm} |p{1.2cm} |p{1.2cm} |}
    \hline
        \textbf{Expt.} & \textbf{Cross Attention Units} & \textbf{Epochs} &  \textbf{Activation} & \textbf{Learning Rate} & \textbf{Sequence Length} & \textbf{Relative Error}\\
        \hline
        \multicolumn{7}{|c|}{\textbf{Advection Equation}}\\
        \hline
        $1$ & $1$ & $40000$ &\texttt{swish} & $5e-5$ & $40$ & $8\%$\\
        \hline
        $2$ & $1$ & $40000$ & \texttt{swish} & $1e-5$ & $40$ & $7.57\%$\\
        \hline
        $3$ & $1$ & $40000$ &\texttt{tanh} & $1e-4$  & $40$ & $7.98\%$\\
        \hline
        $4$ & $1$ & $40000$&\texttt{tanh} & $5e-5$ & $40$ & $8.43\%$\\
        \hline
        $5$ & $1$ & $40000$&\texttt{tanh} & $1e-5$ & $40$ & $7.21\%$ \\
        \hline
        $6$ & $1$ & $40000$&\texttt{tanh} & $1e-5$ & $60$ & $2.61\%$ \\
        \hline
        $7$ & $1$ & $40000$&\texttt{tanh} & $1e-5$ & $80$ & $2.534\%$ \\
        \hline
        $8$ & $2$ & $20000$&\texttt{swish} & $1e-5$ & $40$ & $4.88\%$\\
        \hline
        $9$ & $2$ & $20000$& \texttt{tanh} & $1e-5$ & $60$ & $2.47\%$\\
        \hline 
        \multicolumn{7}{|c|}{\textbf{Burgers Equation}}\\
        \hline
        $1$ & $2$ &$20000$& \texttt{tanh} & $1e-3$ & $40$ & $6.06\%$\\
        \hline
         $2$ & $3$ &$20000$& \texttt{tanh} & $1e-3$ & $40$ & $5.58\%$\\
        \hline
         \multirow{4}{*}{} & & &  & Exponential Decay &  & \\
         & & & &\texttt{learning\_rate}=$1e-3$& & \\
        $3$ & $3$ & $20000$&\texttt{tanh} & \texttt{decay\_rate} = $0.9$ & $40$&$5.24\%$\\
         & & & & \texttt{decay\_steps} = $10000$ & &\\
        \hline
    \end{tabular}
\end{table}

\begin{figure}[!ht]
\centering
\includegraphics[width=1\columnwidth]{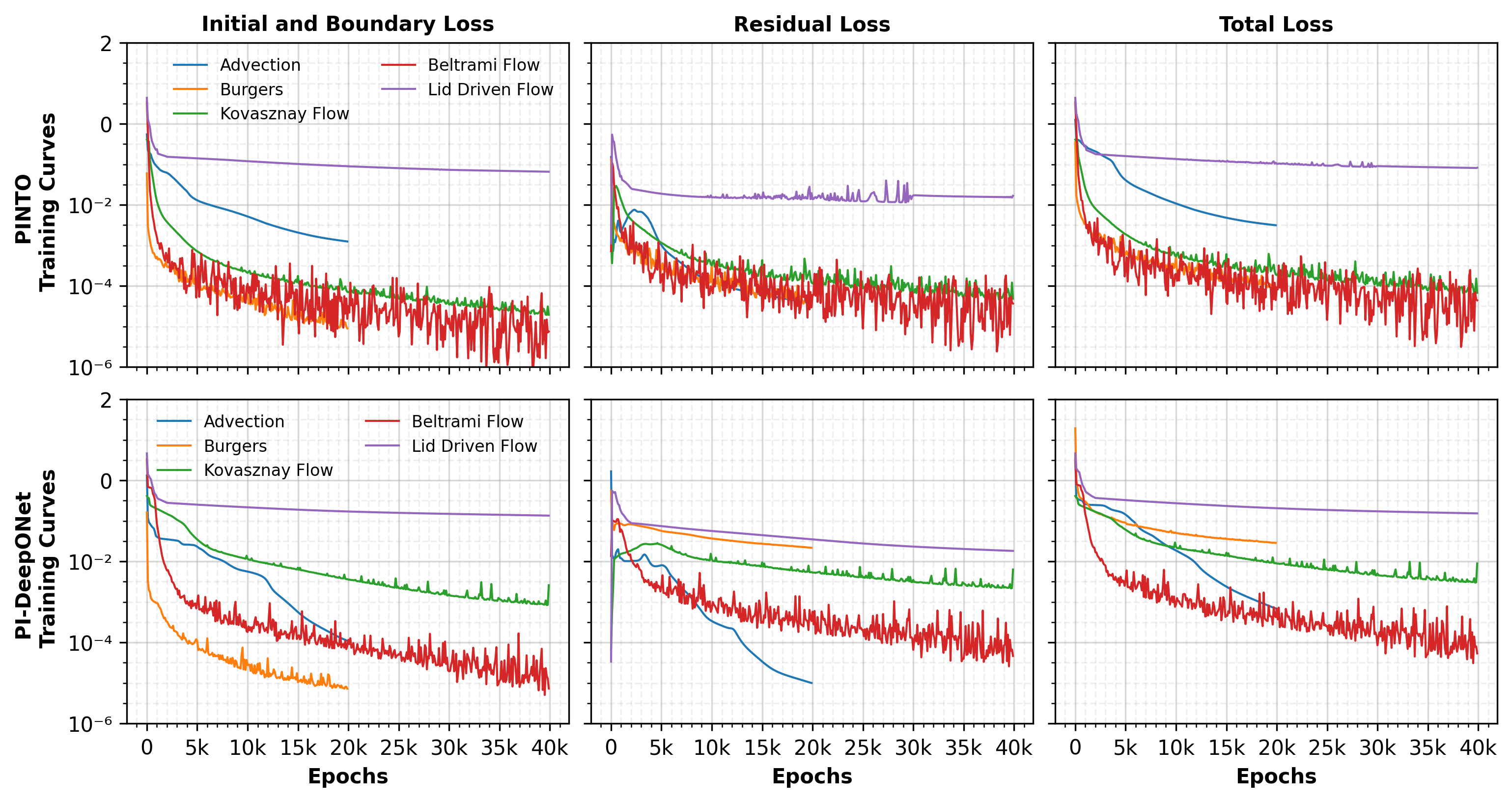} 
\caption{Loss curves during training of PINTO (top row) and PI-DeepONet (bottom row) models for all test cases.}
\label{fig: training_curves}
\end{figure}

\textbf{Hyperparameters and Learning Curves}
Learning curves during training of PINTO and PI-DeepONet for the five test cases are shown in Figure~\ref{fig: training_curves}. For a fair comparison between the PINTO and PI-DeepONet solutions, the models are built approximately with the same number of trainable parameters and have used similar training strategies to ensure their convergence. 
Table~\ref{tab: PINTO-hyperparameter} gives information on the architecture of PINTO models. The hyperparameters required to train the PINTO and PI-DeepONet models are tabulated in Table.~\ref{tab: Training-hyperparameter}.
\begin{table}[!ht]
    \centering
    \caption{Hyperparameters in building PINTO model for each numerical example.}
    \label{tab: PINTO-hyperparameter}
    \begin{tabular}{| p{1.4cm} | p{1cm} | p{1.5cm} | p{0.9cm} | p{0.7cm} | p{0.8cm} | p{1.1cm} | p{0.8cm} | p{0.9cm} | p{0.9cm} |}
    \hline
        \multirow{2}{*}{} & \multicolumn{2}{|c|}{\textbf{\# Parameters}} & \multicolumn{2}{|p{1.6cm}|}{\textbf{QPE, BPE, BVE}} &  \multicolumn{4}{|c|}{\textbf{Cross-Attention Unit}} & {\textbf{Ouput}}\\
        \hline
        \textbf{Test Case}& \textbf{PINTO} & \textbf{PI} & \textbf{Layers} & \textbf{Units} & \multicolumn{2}{|c|}{\textbf{MHA}} & \textbf{\#} & \textbf{Layers}& \textbf{Layers}\\
         & & \textbf{DeepONets} &  & & \texttt{heads} & \texttt{key\_dim} & \textbf{CAUs} &\textbf{(Units)} &\textbf{(Units)}\\
        \hline
        Advection Equation & $100289$ & $109400$ & $2$ & $64$ & $2$ & $64$ & $2$ & $2(64)$ & $2(64)$\\
        \hline
        Burgers Equation & $141825$ & $208896$ & $2$ & $64$ & $2$ & $64$ & $3$ & $2(64)$ & $2(64)$\\
        \hline
        Kovasznay Flow & $75779$ & $69568$ & $2$ & $64$ & $2$ & $64$ & $1$ & $1(64)$ & $2(64)$\\
        \hline
        Beltrami Flow & $75779$ & $69568$ & $2$ & $64$ & $2$ & $64$ & $1$ & $1(64)$ & $2(64)$\\
        \hline
        Lid Driven Cavity Flow & $112834$ & $91264$ & $2$ & $64$ & $2$ & $64$ & $1$ & $2(64)$ & $2(64)$\\
        \hline
    \end{tabular}
\end{table}
\begin{table}[!ht]
    \centering
    \caption{Hyperparameters required to train the PINTO and PI-DeepONet model for all considered numerical examples.}
    \label{tab: Training-hyperparameter}
    \begin{tabular}{| p{1.3cm} | p{0.9cm} | p{1cm} | p{1cm} | p{0.8cm} | p{1.2cm} | p{3cm} | p{0.9cm} | p{1.5cm} |}
    \hline
        \multirow{2}{*}{} & & & & \multicolumn{2}{|p{1.5cm}|}{\textbf{Optimizer}} & \textbf{LR Scheduler} & \multicolumn{2}{|c|}{\textbf{Sequence Length}}\\
        \hline
        \textbf{Test Case}& \textbf{Epochs} & \textbf{Domain Points} & \textbf{Num. Batches} & \textbf{Type} & \textbf{Learning Rate} & & \textbf{PINTO} & \textbf{PI- DeepONets} \\
        \hline
         Advection Equation & $20000$ & $2000$ & $10$ & \texttt{Adam} & $1e-5$ & - & $60$ & $80$ \\
        \hline
         Burgers Equation & $20000$ & $2000$ & $6$ &
        \texttt{Adam} & $1e-3$  & \texttt{Exponential} & & \\
        &  & & & & & \texttt{rate}=$0.9$ & $40$ & $80$\\
        &  & & & & & \texttt{steps}=$10000$ & & \\
        \hline
        Kovasznay Flow & $40000$ & $2000$ & $5$ &
        \texttt{Adam} & $5e-4$ & - & \multicolumn{2}{|c|}{$80$} \\
        \hline
        Beltrami Flow & $40000$ & $5000$ & $5$ &
        \texttt{Adam} &  $1e-4$ & - & \multicolumn{2}{|c|}{$100$} \\
        \hline
        Lid Driven Cavity Flow & $50000$ & $5000$ & $5$ & \texttt{AdamW} & $1e-3$ &\texttt{PiecewiseConstant} & \multicolumn{2}{|c|}{}\\
         & & & & & & \texttt{boundaries }$\left[5000,\,10000\right]$ & \multicolumn{2}{|c|}{$40$}\\
         & & & & & & \texttt{values }$\left[1e-3, 1e-4, 1e-5\right]$ & \multicolumn{2}{|c|}{} \\
        \hline
    \end{tabular}
\end{table}
\section{Computational Time} \label{app: computational-time}
The Table~\ref{tab:computational_time} provides the inference time for each test case. The inference time is reported as the mean of $10$ individual forward passes performed on \texttt{Nvidia 48GB RTX A6000} single GPU machine.
\begin{table}[!ht]
    \centering
    \caption{Training and inference time}\label{tab:computational_time}
    \begin{tabular}{|c|c|c|}
    \hline
        \textbf{Test Case} & \textbf{Inference Time} & \textbf{Grid Size}\\
        & in \textbf{milli seconds} (ms) & \\
        \hline
        1D Advection equation & $40$ms & $1024 \times 100\,\left(x, t\right)$ \\
        \hline
        1D Burgers equation & $35$ms & $1024 \times 100\,\left(x, t\right)$\\
        \hline
        Kovasznay Flow & $185$ms & $256 \times 256\,\left(x, y\right)$\\
        \hline
        Beltrami Flow & $275$ms & $64 \times 64 \times 20 \,\left(x, y, t\right)$\\
        \hline
        Lid Driven Cavity Flow &$100$ms & $256 \times 256\,\left(x, y\right)$\\
        \hline
    \end{tabular}
\end{table}
\section{Interpretation from FDM} \label{app: interpretation_FDM}
Consider the implicit discretization scheme with a forward difference in time and a center difference in space for the 1D advection equation (Eq.~\ref{eq: Adv_eq}) with Dirichlet boundary conditions, and the expression to compute the solution at $\left(n+1\right)^{th}$ step from $n^{th}$ time step for spatial domain with $4-$ grid points is as follows,
\begin{equation*}\begin{bmatrix}
u_0^{n+1} \\
u_1^{n+1} \\
u_2^{n+1} \\
u_3^{n+1}
\end{bmatrix} =
\begin{bmatrix}
1 & 0 & 0 & 0 \\
\frac{\psi}{1+\psi^{2}} & \frac{1}{1+\psi^{2}} & 0 & \frac{\psi^{2}}{1+\psi^2} \\
\frac{\psi^{2}}{1+\psi^2} & \frac{\psi}{1+\psi^{2}} & \frac{1}{1+\psi^{2}} & \frac{-\psi}{1+\psi^{2}} \\
0 & 0 & 0 & 1\\
\end{bmatrix}
\begin{bmatrix}
u_0^n \\
u_1^n \\
u_2^n \\
u_3^n \\
\end{bmatrix}.
\end{equation*}
Here, $\psi=\frac{\beta\Delta t}{2\Delta x}$, $\Delta t$, and $\Delta x$ are discretization lengths in time and space respectively. The solution in the $\left(n+1\right)^{th}$ step of a grid point is a weighted sum of all grid points at $n^{th}$ time step; iteratively, it is a weighted sum of initial values and the weights are a function of $\Delta x$ and $\Delta t$. Now, consider the attention score $\zeta_{i}$ from eq.~\ref{eq: attention_score}, neglecting the normalizing constant and replacing $A_{h}\mu_{t}\left(X\right)$ with $q$, and $B_{h}MLP_{bpe}\left(X_{b}^{i}\right)$ with $k_{i}$ the resulting expression will be $\exp\left(\frac{q^{T}k_{i}}{\sqrt{m}}\right)$, further it is decomposed into two terms \citep{song2021implicit} as follows: 
\begin{equation*}
    \exp\left(\frac{-||q-k_{i}||^{2}_{2}}{2\sqrt{m}}\right)\,\times \exp\left(\frac{||q||^{2}_{2} + ||k_{i}||^{2}_{2}}{2\sqrt{m}}\right),
\end{equation*}
the first term is the RBF kernel with fixed length scale hyperparameter, $\sqrt[4]{m}$ that measures the similarity score between query and key using squared Euclidean distance $||q-k_{i}||^{2}_{2}$, and the second term is the magnitude. The computation of attention scores has a nonlinear relationship with the distance between query and key, similar to the weights in the discretization scheme discussed above, which are functions of discretization lengths.
\section{Physics-Guided training for Burger's equation test case} \label{app: data-driven-training}
\begin{figure}[!ht]
\centering
\includegraphics[width=0.8\columnwidth]{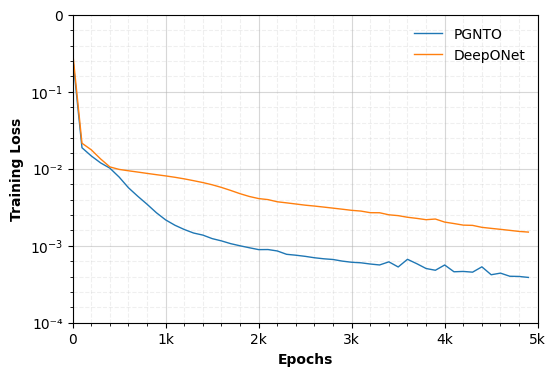} 
\caption{Learning curves of PINTO and PI-DeepONets model for data-driven training of Burgers equation.}
\label{fig: data_driven_training_curves}
\end{figure}

In this Appendix, we present results from simulation data-driven training (physics-guided neural transformer operator - PGNTO) of our architecture for the Burger's equation test case. Here, we don't use physics loss and compare results with the DeepONet. Table~\ref{tab: data-driven-training} shows that the PINTO model performs better than the DeepONet. Figure~\ref{fig: data_driven_training_curves} shows the learning curves of PINTO and DeepONet during training.
 
\begin{table}[!ht]
    \centering
    \caption{Hyperparameters and results of physics-guided training for the Burger's equation test case}
    \label{tab: data-driven-training}
    \begin{tabular}{|c|c|c|c|c|c|c|c|}
        \hline
        \textbf{Model} & \textbf{Epochs} & \textbf{Optimizer} & \textbf{Learning} & \textbf{Trainable} & \textbf{Training} & \multicolumn{2}{|c|}{\textbf{Performace Metrics}} \\
        & & & \textbf{Rate} & \textbf{Parameters} & \textbf{Time} & \textbf{Training} & \textbf{Testing}\\
        \hline
        \textbf{PGNTO} & $5000$ & \texttt{Adam} & $1e-4$ & $141825$ & $75$mins & $0.691\%\,\left(0.646\%\right)$ & $1.26\%\,\left(1.52\%\right)$\\
        \hline
        \textbf{DeepONets} & $5000$ & \texttt{Adam} & $1e-4$ & $140288$ & $5$mins & $1.81\%\,\left(1.874\%\right)$ & $3.01\%\,\left(2.82\%\right)$\\
        \hline
    \end{tabular}
\end{table}

\section*{Data Availability}
All data and code used and developed in the current work have been made publicly available through GitHub and Zenodo \cite{b_sumanth_kumar_2024_14330861}.

\bibliographystyle{elsarticle-num}

\end{document}